# THE LIMITS OF BPE TOKENIZATION IN POLISH: SEGMENTATION-FLEXIONAL FORMS, GRAMMATICAL ANCHORING, AND FIRST-PERSON STABILITY IN INFLECTIONAL LANGUAGE MODELS

**Elżbieta Dawidek** University of Lower Silesia DSW, Wrocław, Poland

**Abstract**

This article analyzes BPE tokenization in Polish as a test case for a broader problem: the limits of statistical segmentation in an inflectional language. The starting point is the question of whether, and to what extent, frequency-based tokenization interacts with or preserves units that are relevant for the linguistic description of Polish: orthographic form, the segmentation-phonemic level, syllabic division, logotomes, prefixes, derivational bases, inflectional endings, grammatical form, and the position of the speaking subject.

The empirical material includes the tokenization of diagnostic words, a simple children's text, selected forms from the Preamble to the Constitution of the Republic of Poland, word-family tests, and examples containing Polish diacritics and nasal vowels. The analysis shows that BPE tokenizers may locally produce segments that coincide with syllabic, logotomic, or morphologically interpretable divisions, but they remain fundamentally dependent on the frequency of written forms. They do not systematically move from orthographic representation to the segmentation-phonemic level or to context-dependent phonetic realization.

The results also indicate that BPE stabilizes not grammatical categories themselves, but frequent surface fragments of their exponents. Segments such as *-y*, *-ich*, *-ni*, or *-ć* may appear as stable tokens or parts of tokens, but this does not amount to preserving the full grammatical form. A form such as *ustanawiamy* is not merely a sequence ending in *-y*; it is a verbal form anchored in conjugation, person, number, tense, mood, and aspect.

The article develops the concept of grammatical form anchoring, arguing that in Polish a form may not only encode syntactic relations, but also establish the position of the speaking subject. In forms such as *poszłam*, *zrobiłam*, or *byłam*, the grammatical "I" need not be expressed by a separate pronoun, because it is anchored in inflection. In interaction with AI, this reveals an additional problem: a language model does not possess its own stable grammatical "I", but reconstructs it contextually, often mirroring the user's forms or shifting grammatical gender.

I propose treating Rocławski's segmentation-flexional forms not as a description of how BPE operates, but as a diagnostic framework for assessing the quality of tokenization boundaries. The findings lead to the research hypothesis that more stable modeling of Polish may require several layers: sublexical stabilization inspired by Rocławski's system, anchoring grammatical form in the inflectional system, representing sentence patterns and verbal valency structure, and maintaining the position of the grammatical "I" in dialogue. From this perspective, the problem of Polish AI is not reducible to reducing the number of tokens. Rather, it concerns maintaining relations between orthographic representation, word form, grammatical function, sentence position, and the position of the speaking subject.

**Keywords**

BPE tokenization; Polish; Rocławski's segmentation-flexional forms; logotomes; orthographic representation; segmentation-phonemic level; grammatical form; grammatical form anchoring; grammatical identity; grammatical "I"; inflection; language models; AI; valency; sentence patterns

## Introduction

Tokenization is one of the first stages through which text is processed by large language models. In the Byte Pair Encoding (BPE) algorithm, text is divided into units determined by the statistical frequency of co-occurring character sequences in the training corpus. From a technical perspective, this is a procedure of compression and segmentation. From the perspective of an inflectional language, however, it is not neutral.

In Polish, a large part of grammatical information is located inside the word. A word form may simultaneously carry information about person, number, gender, case, tense, mood, aspect, and syntactic relation. For this reason, the problem of Polish tokenization is not limited to the number of tokens per word. The more important question is whether token boundaries preserve the structure needed to reconstruct the grammatical form and its place in the sentence.

This article approaches BPE tokenization through several levels of linguistic description relevant to Polish: orthographic representation, phonemic-didactic segmentation, phonetic realization, morphology, inflection, syntax, and the position of the speaking subject. A particularly important point of reference is Bronisław Rocławski's theory of segmentation-flexional forms, which makes it possible to capture the non-transparent relations between letters, segmental units, sounds/phonemes, syllables, and logotomes.

The data analyzed in this study show, however, that local correspondence with sounds/phonemes in a didactic sense, with syllables, or with logotomes is not sufficient for assessing the quality of tokenization. A tokenizer may stabilize segments that coincide with units used in linguistic description, while still failing to preserve the anchoring of grammatical form. The problem of Polish tokenization, therefore, is not simply that BPE "divides words incorrectly", but that it operates at the graphemic-frequency level, whereas Polish requires the preservation of relations between orthographic representation, phonemic-didactic segmentation, phonetic realization, inflectional form, sentence structure, and the position of the speaking subject.

A segment such as *-y*, *-ich*, *-ni*, or *-ć* may be frequent and statistically stable, but it does not necessarily represent the full grammatical function it performs in the inflectional system. The form *ustanawiamy* is not simply a string ending in *-y*; it is a verbal form anchored in conjugation, person, number, tense, mood, and aspect. Similarly, *ogólnoludzkich* is not merely a sequence ending in *-ich*, but an adjectival form anchored in declension, number, gender, and case.

The problem of grammatical form anchoring has an even deeper dimension. In Polish, a form may not only encode a grammatical category, but also establish the position of the speaking subject. In utterances such as *poszłam*, *zrobiłam*, *byłam*, or *pomyślałam*, the pronoun *ja* 'I' need not appear at all, and yet the subject is anchored in the verbal form. The form indicates who is speaking, from what position they are speaking, and how they are inscribed into the structure of the sentence.

In interaction with a language model, this problem becomes particularly visible. AI does not possess its own stable grammatical position in the world. It has no body, no gender, no experience, and no "I" anchored in inflection. When generating an utterance in Polish, however, it must choose a form: *zrobiłem* or *zrobiłam*, *byłem* or *byłam*, *gotowy* or *gotowa*. In English, this problem is less visible, because the first person does not require such consistent gender agreement. In Polish, it appears immediately.

The main thesis of this article is therefore as follows: BPE tokenization in Polish stabilizes frequent fragments of written form, but it does not guarantee the stabilization of grammatical form or of the grammatical "I". For an inflectional language, this means that a multilayered analysis is needed: sublexical stabilization inspired by Rocławski's segmentation-flexional system, anchoring grammatical form in the system of conjugation and declension, representing sentence patterns, and maintaining continuity in the position of the subject in dialogue.

The article addresses five research questions:

1. Does BPE tokenization in Polish move from the written string to the segmentation-phonemic or phonetic-realizational level?
2. Are token boundaries aligned with syllabic, logotomic, or morphologically interpretable divisions?
3. Does BPE stabilize prefixes, derivational bases, inflectional endings, and grammatical exponents as units of the language system, or only as frequent fragments of written form?
4. Is a stable token equivalent to the anchoring of grammatical form?
5. What happens to the grammatical "I" in Polish when an utterance is generated by a language model that lacks its own stable subject position?

The working hypothesis is that BPE in Polish does not systematically represent sounds/phonemes, logotomes, morphemes, or grammatical forms. It primarily stabilizes frequently occurring fragments of written form. These fragments may locally coincide with linguistically relevant units, but such coincidence does not yet mean that their function in the Polish language system is preserved.

The second part of the hypothesis concerns language models as systems that generate utterances. In Polish, the problem of grammatical anchoring concerns not only the word, but also the subject. A model may generate first-person forms, but it does not possess its own grammatical "I"; its position is contextual, statistical, and often mirror-like. Therefore, a stable Polish language model must maintain not only correct tokens, but also the relations between form, inflection, sentence pattern, and the position of the speaker.

## 1. Theoretical Framework

The problem of BPE tokenization in Polish requires distinguishing several levels of description. In this article, BPE is not compared directly with "the Polish language" as a whole, but with levels that in Polish remain interconnected while not being identical: orthographic representation, phonemic-didactic segmentation, phonetic realization, morpho-inflectional structure, syntax, valency, and the position of the speaking subject (Rocławski, 1976, 2010; Grzegorczykowa et al., 1998; Łuczyński, 2015).

The first level is the graphemic-frequency level. This is the level on which BPE operates: the algorithm stabilizes frequent sequences of written form rather than linguistic units in the phonological, morphological, or grammatical sense. Byte Pair Encoding was originally developed as an algorithm for compressing character sequences and was later adapted in NLP as a method for segmenting subword units (Gage, 1994; Sennrich et al., 2016). A token, therefore, is not by definition a sound, syllable, morpheme, logotome, or word. It may locally coincide with any of these, but such coincidence results from the frequency of written form, not from the tokenizer implementing linguistic analysis (Bostrom & Durrett, 2020; Gutierrez-Vasques et al., 2023).

The second level is the orthographic level, that is, the written form of the word in the Polish writing system. This level includes letters, digraphs, diacritics, and historically stabilized relations between spelling and pronunciation. For BPE tokenization, this is the primary level, because the algorithm receives written text, not a direct representation of phonological or phonetic structure.

The third level is the segmentation-phonemic level, where I situate Rocławski's theory. Rocławski is not treated here as a description of how BPE operates, nor as a simple representation of actual pronunciation, but as a diagnostic framework that makes it possible to distinguish between the letter, the segmental unit, the phoneme/sound in a didactic sense, the syllable, the logotome, and sites of orthographic difficulty (Rocławski, 1976, 2010).

The fourth level is the phonetic-realizational level, that is, actual or normatively described pronunciation dependent on context. This level includes, among other phenomena, devoicing, voicing assimilation, the realization of nasal vowels, and variation in the pronunciation of consonant clusters. It must be distinguished both from orthographic representation and from the didactic segmentation-phonemic level.

The fifth level is the morpho-inflectional level. It includes prefixes, derivational bases, inflectional endings, conjugation, declension, and the anchoring of grammatical form in the inflectional system. In Polish, a word form may simultaneously carry information about person, number, gender, case, tense, mood, and aspect. For this reason, a token boundary that cuts through a form may have interpretive consequences, not merely technical ones (Grzegorczykowa et al., 1998; Laskowski, 1998; Łuczyński, 2015).

The sixth level is the syntactic-valency level, at which grammatical form acquires its full function only within the structure of the sentence and in relation to the verb as the center organizing the participants of an event. Sentence patterns and valency make it possible to capture relations that cannot be reduced to the mere presence of an inflectional ending (Grzegorczykowa, 1996, 2007; Fillmore, 1968; Tesnière, 1959).

The seventh level is the subject-dialogical level, that is, the grammatical "I". In Polish, the speaking person may be anchored in the verbal form even without the separate pronoun *ja* 'I'. This level is not part of the empirical tokenization analysis in Groups A–E, but constitutes an interpretive extension of the problem of grammatical form anchoring in dialogue (Grabias, 2019; Łuczyński, 2015; Benveniste, 1971; Silverstein, 1976).

Only by separating these levels can we precisely identify what BPE does not do. BPE does not "make a phonetic error", because it does not operate at the phonetic level. It stabilizes written form. The problem is that in Polish, written form is not a sufficient carrier of all phonological, phonetic, inflectional, syntactic, and subject-related relations.

### 1.1. BPE as Statistical, Not Linguistic, Segmentation

Byte Pair Encoding is a segmentation algorithm based on the frequency of co-occurring character sequences. Historically, BPE was developed as a method of data compression, and in natural language processing it was adapted as a way of representing rare and complex forms through subword units (Gage, 1994; Sennrich et al., 2016). In simplified terms, the algorithm begins with small units of written form and then iteratively merges those pairs or sequences that occur sufficiently often in the corpus. The result is a token vocabulary optimized for the statistical compression of textual data.

This mechanism is not designed as linguistic analysis. A token is not, by definition, a letter, sound, phoneme, syllable, morpheme, logotome, or word. It may locally coincide with one of these units, but only when a given unit, or its written representation, reaches sufficient frequency stability in the corpus. This means that BPE does not simply follow the letter, but neither does it follow phonology, phonetics, or morphology. It follows the frequency of written form. For this reason, in this article I refer to the level on which BPE operates as the graphemic-frequency level.

This distinction is particularly important for inflectional languages. In English, which is more strongly positional and has relatively simpler inflectional morphology, frequent character sequences may more often coincide with whole words, stems, or affixes. This does not mean that BPE is morphologically aware; it only means that the structure of the language, the frequency of written form, and compression optimization more often work in the same direction. In Polish, the situation is different: grammatical information is encoded to a much greater extent inside the word form. An inflectional ending, prefix, stem, verbal aspect, or noun case may carry relations that cannot be reconstructed solely from the word's position in the sentence (Grzegorczykowa et al., 1998; Laskowski, 1998; Łuczyński, 2015).

For this reason, subword tokenization is not typologically neutral. Research on multilingual tokenizers shows that languages differ in tokenization cost: the number of tokens per word, the number of bytes per unit of representation, and the losses resulting from a mismatch between the tokenizer and the morphology of the language (Rust et al., 2021; Ahia et al., 2023; Arnett & Bergen, 2025). Bostrom and Durrett (2020) argue that BPE may be suboptimal for language-model pretraining, among other reasons because its boundaries do not necessarily coincide with morphologically interpretable units. Similar concerns motivate work on morphology-aware tokenizers, such as MorphBPE, as well as studies of morphology-aware subword tokenization for Slavic and other morphologically rich languages (Asgari et al., 2025; Držík, 2024; Držík & Kapusta, 2026).

Polish is a particularly clear case of this problem. It is an inflectional language, with an extensive declensional system, conjugation, verbal aspect, and relatively free word order. This means that grammatical relations are often distributed inside the word form and across forms in the sentence. If tokenization divides a word according to the frequency of written form, it may cut through a fragment that is inflectionally relevant. Conversely, it may stabilize a final character sequence that looks like a grammatical exponent but does not preserve the full function of the form.

Two issues should therefore be distinguished: tokenization efficiency and linguistic adequacy of segmentation. A tokenizer better adapted to Polish may reduce the number of tokens required to represent a text. Work on Polish models and tokenizers, including Bielik/APT4 and PLLuM, shows that optimizing a tokenizer for Polish has measurable importance for the efficiency of textual representation (Ociepa et al., 2025, 2026; Kocoń et al., 2025). This does not automatically mean, however, that tokenization preserves grammatical form, phonological structure, or sentence-level relations. A smaller number of tokens is computationally important, but it is not equivalent to the anchoring of grammatical form.

In this sense, the analysis of BPE in Polish cannot be reduced to the question of how many tokens are assigned to a word. The more important question is what is preserved and what is cut by token boundaries. If BPE stabilizes a fragment such as *-y*, *-ich*, *-ni*, or *-ć*, it may preserve a statistical trace of grammar. This does not mean, however, that it preserves grammatical form as a unit of the language system. Similarly, if a token coincides with a syllable or logotome, this does not yet mean that the model has recognized phonological or logotomic structure; it may be a local coincidence resulting from the frequency of written form.

For this reason, I treat BPE in this article as a mechanism of statistical approximation, not as a language analyzer. Its outputs may be compared with levels of Polish linguistic description — orthographic, segmentation-phonemic, phonetic-realizational, morpho-inflectional, and syntactic-valency levels — but they cannot be identified with them. This distinction is crucial for the analysis that follows: a token may be efficient, stable, and frequent, while still failing to anchor grammatical form.

### 1.2. Rocławski: Letter, Segmental Unit, Phoneme/Sound, Syllable, and Logotome

Bronisław Rocławski's theory is treated in this article not as a description of the BPE mechanism and not as a simple representation of actual pronunciation, but as a diagnostic framework for assessing the quality of segmentation. Rocławski makes it possible to distinguish several levels of word divisibility: the letter level, the segmentation-phonemic level, the syllabic level, and the logotomic level. Each of these levels describes a different type of relation between written form and the structure of language (Rocławski, 1976, 2010).

The letter level refers to the graphic signs used in the written form of a word. In Polish, a letter is not the same as a sound or a phoneme. One phoneme may be represented by different spellings, and the same written form may require different interpretation depending on context. For this reason, letter-based segmentation cannot be equated either with phonological analysis or with phonetic realization.

The segmentation-phonemic level in Rocławski's approach makes it possible to identify where written form ceases to be transparent in relation to units relevant for mastering the structure of the word. This has didactic and diagnostic significance: it helps identify places where the reader, the child learning to read, or a system processing text must move from the surface of written form to linguistic structure. In this sense, Rocławski provides not so much a phonetic transcription of pronunciation as a tool for recognizing relations between written form, phoneme/sound, syllable, and larger word components.

The phonetic-realizational level must be distinguished from this approach. It concerns concrete or normatively described pronunciation dependent on context: final devoicing, voicing before a voiced consonant, the realization of nasal vowels, or variation in the pronunciation of consonant clusters. In this article, realizational examples are used not to equate Rocławski's theory with phonetic transcription, but to show that BPE remains at the level of written form and does not systematically move either to the segmentation-phonemic level or to the phonetic-realizational level.

Syllabic division is based on syllable structure and its vocalic nucleus. It may locally coincide with token boundaries, but such coincidence does not yet prove agreement with other levels of description. For example, the segmentation ław|ka may appear to coincide with the syllabic division ław-ka, but it does not reflect phonetic realization, in which w is devoiced to f before k. Coincidence with one level of description may therefore mask divergence at another level. Syllabification is itself convention-dependent: the same string may be divided phonologically (tru-skaw-ka) or as a sequence of logotomes (trsk-awk-a), depending on what is being divided and for what purpose. The syllable therefore functions here as one comparative grid among several, not as a target of tokenization.

Logotomes are functional word components described by Rocławski as units important for mastering word structure in the process of reading and writing. They are not identical with letters, sounds/phonemes, syllables, or morphemes. For the analysis of BPE, they are particularly interesting because statistical tokenization may locally produce segments that coincide with logotomes. Such coincidence does not mean that the tokenizer implements the theory of logotomes. It only means that a certain fragment of written form has achieved sufficient frequency stability in the corpus to become fixed as a token.

Rocławski's distinction between basic and non-basic letters is especially important. A non-basic letter marks a place where orthographic representation is not transparent in relation to the phonemic unit or pronunciation. In this sense, the system of basic and non-basic letters allows one to describe sites of orthographic difficulty precisely: those places where spelling does not directly coincide with an intuitive phonemic or phonetic reading (Rocławski, 1976, 2010).

This distinction is directly relevant to the analysis of BPE. A tokenizer operates on written text and therefore stabilizes surface configurations of orthography. If Polish spelling preserves historical or systemic divergences between letter, phoneme, and pronunciation, BPE may stabilize precisely such an orthographic difficulty as a frequent graphemic fragment. This does not mean that the model recognizes the phonological relation behind the written form.

An example is the pair *morze/może*. Both words may be reduced to the same segmentation-phonemic level: *m-o-ż-e*. They differ, however, in their letter sequence: *morze* contains the digraph *rz*, whereas *może* contains *ż*. A tokenization that differentiates these forms therefore does not show a phonological difference, but a graphemic-frequency difference. BPE stabilizes spelling, not phonemic equivalence.

Another type of example is represented by forms such as *lód* and *nóż*. In these words, the issue is not a "change from *ó* to *u*" as a phonetic process, but an orthographic-phonemic relation: *ó* is the written representation of the sound/phoneme /u/. Only additionally, at the realizational level, final devoicing appears: *d* becomes *t* in *lód*, and *ż* becomes *sz* in *nóż*. Such examples show that the analysis of tokenization must separate orthographic representation, the segmentation-phonemic level, and phonetic realization.

Words with nasal vowels, such as *ząb*, *ręka*, *kąt*, or *kąpiel*, show the necessity of distinguishing the simplified segmentation-phonemic level from the phonetic-realizational level. In spelling, *ą* and *ę* occur as single graphemes, whereas in pronunciation their realization depends on phonetic context. The example *ząb* may therefore be described on different levels: orthographically as *ząb*, segmentationally with *ą* preserved as a unit of spelling/phoneme, and realizationally with the decomposition of nasality before a labial consonant and final devoicing of *b* to *p*. BPE, however, primarily stabilizes orthographic representation.

Rocławski's table of phoneme and phoneme-diad frequencies is of particular importance for this article. Rocławski measured the statistical regularity of co-occurring phonemic units as a structural property of Polish. BPE also operates on frequency of co-occurrence, but on different material: sequences of written form in a textual corpus. There is therefore partial methodological compatibility between these approaches, but not epistemological identity. Rocławski describes the structure of language through phonemic and didactic data; BPE extracts regularities of written form as an effect of statistical compression.

For this reason, I do not claim in this article that BPE "discovers" logotomes, phonemes, or Rocławski's system. I make a more cautious claim: token boundaries may be compared with the levels described by Rocławski because both approaches touch on segmentation regularities. If a token coincides with a logotome, a syllable, or a morphologically interpretable fragment, this is a local convergence that requires interpretation. It is not evidence that the tokenizer possesses a phonological or didactic representation in Rocławski's sense.

In the shortest terms: BPE does not tokenize the phonology of Polish. It tokenizes its written form — together with historically and systemically stabilized orthographic difficulties. Rocławski's theory, by contrast, allows us to name and organize the places where written form, segmental structure, and phonetic realization diverge.

**Note on the syllable.** Across this series, the syllable is not used as a unit of tokenization or as a benchmark target. It is not a fixed division unit: its boundaries depend on the segmentation convention adopted (phonological syllabification tru-skaw-ka vs. logotome segmentation trsk-awk-a). The proposed metrics and diagnostic levels are defined over morpheme boundaries, with the syllabic and logotomic levels serving as comparative grids only.

### 1.3. Grammatical Form in Polish: Conjugation, Declension, and Inflectional Exponents

In Polish, grammatical form is not an accidental sequence of final letters. It is a unit anchored in the inflectional system: in conjugation or declension, in person, number, gender, case, tense, mood, and aspect. These categories are described systematically in the Polish grammatical tradition and cannot be reduced to the surface frequency of final fragments of written form (Grzegorczykowa et al., 1998; Laskowski, 1998; Łuczyński, 2015; Wróbel, 2001).

For this reason, segmenting a word into tokens may have not only technical, but also interpretive consequences. If a token boundary cuts through an element of a word form, it does not merely cut through a sequence of characters. It may cut through part of an inflectional exponent, a prefix, a derivational base, or a component of the form that participates, within the language system, in marking person, number, gender, case, tense, mood, or aspect. What from the perspective of BPE is only a frequent graphemic sequence may, from the perspective of inflection, be part of a grammatical relation.

For example, the form *ustanawiamy* is not simply a character sequence ending in *-y*. It is a verbal form that carries information about the first person plural, present tense, indicative mood, and imperfective aspect. If a tokenizer segments it as *ustan|aw|iam|y* or *u|stan|aw|iam|y*, it stabilizes the final *-y* as a frequent fragment of written form, but it does not preserve the full personal ending or the full anchoring of grammatical form. This means that a token may locally coincide with part of an inflectional exponent while still failing to carry the complete grammatical information of the form.

Similarly, *ogólnoludzkich* is not merely a word ending in *-ich*. It is an adjectival form anchored in declension, number, gender, and case. A tokenization such as *ogól|no|lud|zk|ich* or *og|ó|ln|ol|ud|zk|ich* may stabilize final *-ich* as a frequent surface fragment, but it does not necessarily represent its grammatical function. In such a case, the token preserves a fragment of the exponent, but it does not guarantee preservation of the form's relation to the inflectional paradigm.

This distinction makes it possible to separate three levels: a fragment of written form, a grammatical exponent, and an anchored grammatical form. A fragment of written form may be statistically stable. A grammatical exponent may be partially visible in the token. An anchored grammatical form, however, requires reference to the inflectional system and to the relations that the form activates in the sentence. Therefore, a frequent fragment is not the same as a morpheme, and a stable token is not the same as an anchored grammatical form.

The analysis therefore points to a difference between a statistical exponent and a grammatical form. BPE may stabilize part of an inflectional exponent, but this does not automatically mean that it stabilizes the form within the inflectional system. This is especially important for Polish, because prefixes, suffixes, inflectional endings, stem alternations, and aspectual exponents are described systemically. They depend on conjugational and declensional patterns, not only on the frequency of surface character sequences (Grzegorczykowa et al., 1998; Laskowski, 1998; Łuczyński, 2015; Saloni et al., 2012).

This point can be further clarified by comparison with morphological tools for Polish. Analyzers such as Morfeusz do not segment text solely according to the frequency of written form, but assign possible morphosyntactic interpretations to forms: lemma, part of speech, and a set of grammatical features (Woliński, 2006, 2014). This does not mean that a tokenizer should operate like a morphological analyzer. It does, however, show the difference between levels: BPE operates on the graphemic-frequency surface of text, whereas inflectional analysis relates a form to a system of grammatical categories.

A further interpretive observation follows from this contrast. Lemmatization and tokenization are functionally inverse procedures: lemmatization reduces an inflected form to its dictionary form, whereas tokenization decomposes a surface form into smaller units. The exact inverse of lemmatization, however, is morphological segmentation — the decomposition of a form into morphemes according to the structure of the language (Goldman & Tsarfaty, 2022). BPE is not this. It determines boundaries by the frequency of written form, not by the morphological system. For this reason I describe statistical tokenization in Polish as a pseudo-inverse of lemmatization: it moves in the same direction as the inverse of lemmatization — from form toward fragments — but it is not that inverse, because its boundaries are graphemic-frequency boundaries, not morpheme boundaries. For a positional language this approximation is largely sufficient; for an inflectional language it is structurally inadequate, because the units it produces need not coincide with the morphemes carrying grammatical information.

In this article, therefore, the concept of grammatical form anchoring refers not to the mere occurrence of an ending or one of its fragments in tokenization, but to the possibility of relating the form to the inflectional system and to the relations that the form maintains in the sentence. Tokenization based solely on frequency of written form may preserve a statistical shadow of grammar, but without its relational function.

### 1.4. Grammatical Form Anchoring and the Grammatical "I"

The concept of grammatical form anchoring should be understood more broadly than as the stabilization of an inflectional ending. In Polish, grammatical form may not only encode categories of inflection, but also constitute the position of the speaking subject. The utterance *poszłam* is a single word and, at the same time, a complete sentence: the subject is not expressed as a separate pronoun, yet it remains present in the verbal form as an implied subject. The form indicates person, number, feminine gender, perfective aspect, and past tense. What in English is expressed by the construction *I went* may in Polish be anchored in the inflection of the verb itself (Łuczyński, 2015; Grabias, 2019).

In this sense, the inflectional ending is not merely an addition to lexical meaning. It is the place where language may anchor the subject. Grammatical form does not only describe an action; it also indicates who is speaking, from what position they speak, and how they are inscribed into the structure of the sentence. In Polish, *ja* 'I' does not always have to be expressed as a separate pronoun. Very often, it becomes visible precisely through inflection.

This problem can also be described from a broader linguistic perspective: personal and gendered forms have an indexical character because they point to the relation between the utterance, the speaker, and the communicative situation. They are therefore not only formal markers, but elements organizing the position of the subject in the speech act (Benveniste, 1971; Silverstein, 1976). In Polish, this relation is particularly visible because first-person past-tense forms require the choice of grammatical gender: *zrobiłam* or *zrobiłem*, *byłam* or *byłem*, *poszłam* or *poszedłem*.

In human–language model interaction, this problem becomes especially evident. A language model has no body, no biographical experience, and no stable grammatical "I" of its own. When generating a first-person utterance in Polish, however, it must choose a form of gender, person, number, and tense on the basis of context. It does not speak from its own place in the grammatical system, but reconstructs that place statistically. It may therefore adopt the user's form, mirror it, or shift between personal genders.

For a user of Polish, such a shift is not neutral. The difference between *zrobiłam* and *zrobiłem*, *byłam* and *byłem*, *poszłam* and *poszedłem* is not merely a stylistic variant. It is a change in the grammatical position of the subject. For this reason, grammatical form anchoring includes not only the stabilization of an inflectional exponent, but also the maintenance of the relation between form, speaking person, and that person's place in the sentence.

In this article, the problem of the grammatical "I" is not treated as a separate group of tokenization data. It therefore does not belong to the same level of analysis as the examples of words, word families, or texts subjected to BPE segmentation. It is a subject-dialogical level: an interpretive extension of the problem of grammatical form anchoring in the context of utterance generation in interaction. Examples such as *poszłam / poszedłem*, *zrobiłam / zrobiłem*, and *byłam / byłem* serve to show that in Polish grammatical form may anchor the position of the speaking subject. Their systematic analysis would require a separate dialogical study, focused on the stability of first-person, gendered, and personal forms in longer human–model interactions.

For this reason, the grammatical "I" appears in the article not as a result of the tokenization of a single word, but as a theoretical consequence. If tokenization and the subsequent layers of the model do not maintain a stable relation between form, inflection, and the position of the speaker, the model may generate utterances that are locally correct in formal terms but unstable at the level of subject position. In Polish, such instability becomes visible more quickly than in languages in which the first person does not require consistent gender agreement.

### 1.5. The Verb as the Center of Relations: Sentence Patterns and Valency

Grammatical form anchoring does not end at the level of the word. In Polish, the full value of a form becomes visible only in the sentence, in relation to other forms. The verb is especially important here, because it organizes the structure of the event and opens positions for participants: who acts, what they do, to whom, with what, where, when, and for what purpose.

Sentence patterns described by Renata Grzegorczykowa, together with the tradition of valency analysis, make it possible to capture this level of relations. The verb is not merely one element of the sentence. It is the center organizing the syntactic-semantic structure: it determines possible participant roles in the event, specifies required or expected argument positions, and links them with appropriate morphological forms. In this sense, the verbal form indicates person, number, tense, mood, and aspect, while also opening specific argument positions (Grzegorczykowa, 2007; Fillmore, 1968; Tesnière, 1959).

This level of analysis is necessary because the mere presence of an inflectional exponent is not sufficient to reconstruct the function of the form. The example *ustanawiamy* shows that tokenization may preserve part of an inflectional exponent, but it does not necessarily preserve the full personal form. Only in a sentence does this form acquire its proper value: *ustanawiamy* something, for someone, for some purpose, within a specific normative or communicative act. The ending -*y* alone is not enough to reconstruct these relations. The form must be anchored in the structure of the sentence.

Similarly, the infinitive *przekazać* is not merely a form ending in *-ć*. In morphological analysis, it may be considered as a structure consisting of the prefix *prze-*, the base *kaza-/kazać*, and the infinitive marker *-ć*. In a sentence, this verb opens positions for a sender, a recipient, and the content being transmitted. If tokenization cuts the prefix, the base, and the grammatical exponent in a way determined by surface frequency, the model may lose stable access to the relations organized by the verb.

For this reason, the analysis of Polish tokenization must go beyond the number of tokens. It requires asking whether the word form remains anchored in the inflectional system, in sentence structure, and in the position of the speaking subject. In Polish, the token must be related to inflection, inflection to the sentence, and the sentence to the position of the speaking "I".

The empirical analysis follows from these assumptions. In this study, tokenization is not evaluated solely in terms of compression efficiency, that is, the number of tokens per word or text. I also analyze the relation between token boundaries and orthographic, segmentation-phonemic, phonetic-realizational, syllabic, logotomic, morphological, grammatical, and sentence-valency patterns. The raw data include token segmentation, token identifiers, and token counts. The interpretation concerns whether tokenization preserves, fragments, or obscures the anchoring of grammatical form.

The problem of the grammatical "I", by contrast, is transferred to the subject-dialogical level. It is not analyzed as a separate group of tokenization data, but as a theoretical consequence: if grammatical form in Polish can anchor the speaker, then the stability of a language model requires not only correct segmentation and inflection, but also the maintenance of relations between form, sentence, and the position of the subject in dialogue.

## 2. Methodology

The aim of the study is to provide a qualitative analysis of BPE tokenization in Polish from the perspective of the correspondence between token boundaries and selected levels of linguistic description: the graphemic-frequency, orthographic, segmentation-phonemic, phonetic-realizational, syllabic, logotomic, morpho-inflectional, and grammatical-sentential levels. The study is exploratory and diagnostic in character. It does not constitute a full benchmark of tokenizers for Polish, but is intended to identify points at which statistical segmentation of written form preserves, fragments, or obscures structures that are important for Polish as an inflectional language.

The basic unit of analysis is a word or short test text. In the case of individual words, both the tokenization output and its relation to several reference levels were analyzed: orthographic representation, the segmentation-phonemic level, syllabic division, possible logotomic division, the phonetic-realizational level, and morpho-inflectional structure. In the case of short texts, the analysis focused on overall tokenization efficiency and on the types of forms that were either stabilized or internally segmented.

A separate interpretive category was grammatical form anchoring. This was not equated with the mere presence of an ending or one of its fragments in tokenization. I assumed that a grammatical form is anchored only when it can be related to the inflectional system: conjugation or declension, person, number, gender, case, tense, mood, or aspect. The analysis also took into account the fact that, in Polish, grammatical form may anchor the speaking "I", especially in first-person and gendered forms. This problem, however, was not treated as a separate group of tokenization data, but as a theoretical and interpretive extension of the problem of grammatical form anchoring.

### 2.1. Models and Tokenizers

The study compared three tokenization environments:

1. **Bielik/APT4 Tokenizer** — the tokenizer available in the Bielik/APT4 environment, treated as an example of tokenization adapted to Polish.
2. **OpenAI-current Tokenizer** — a newer tokenization environment available in the OpenAI interface examined in this study.
3. **OpenAI-legacy Tokenizer** — an older tokenization environment associated with the GPT-4/GPT-3.5 legacy family.

Because of the limited transparency of public documentation concerning the exact version of the tokenizers used in the tested environment, this article uses descriptive labels. **OpenAI-current Tokenizer** refers to the newer observed tokenization environment, **OpenAI-legacy Tokenizer** refers to the older observed tokenization environment, and **Bielik/APT4 Tokenizer** refers to the tokenizer available in the Bielik/APT4 environment. These labels are not intended as official names of architectures, but as a way of preserving the distinction between the three observed modes of segmentation.

For each tested word or text, the following information was recorded:

1. the textual segmentation visible in the tokenization tool;
2. token identifiers, where available;
3. the number of tokens;
4. the number of characters for short texts;
5. technical notes concerning separators, newline characters, and visual interpretation.

In the analysis of token counts, tokens corresponding only to newline characters or technical separators were omitted if they were not part of the word under examination. Textual segmentation in OpenAI tokenizers was read from the color-coded visualization, while token identifiers were recorded separately from the Token IDs panel. In cases where the color boundary of a token was ambiguous, the result was marked as requiring verification.

For a replicable version of the study, it is necessary to provide the exact name of the tool, the library version, the date of testing, the full list of tokens, token identifiers, and the input text in unchanged form. This is important because tokenization results may vary across tokenizer families, model versions, and testing environments.

## 2.2. Research Material

The research material was selected in order to test several different types of relations between written form, word structure, grammatical function, and the possible anchoring of form in the inflectional system. The purpose was not only to compare the number of tokens, but to capture the quality of segmentation boundaries.

The empirical material was divided into five diagnostic groups.

**Group A** included a simple children's text:

*Siała baba mak, nie wiedziała jak. A dziad wiedział nie powiedział, a to było tak.*

This text was used as a high-frequency control material with a simple syntactic structure. It made it possible to test whether textual simplicity, rhythm, and the presence of frequent words lead to the stabilization of whole forms, or whether inflectional forms still undergo internal segmentation.

**Group B** included selected forms from the Preamble to the Constitution of the Republic of Poland and other official or complex forms:

*Rzeczypospolitej, zobowiązani, ustanawiamy, ogólnoludzkich, przekazać, wdzięczni.*

This material was used to evaluate the segmentation of long Polish forms in which syllabic, morphological, inflectional, and derivational elements co-occur.

**Group C** included a word family related to the base *kazać*:

*przekazać, kazać, pokazać, zakazać, zakaz, zakazany, pokazujemy.*

The purpose of this group was to test whether the tokenizer stabilizes a shared base or morphological family, or rather recurring surface fragments such as *aza*, *az*, *zak*, *pok*, *ć*, *any*, *uj*, and *emy*.

**Group D** included phonologically diagnostic words:

*jabłko, chleb, ząb, lód, nóż, lekarz, morze, może, jesień, prośba, ławka.*

This material tested the relation between the letter sequence and several reference levels: orthographic representation, phonemic-didactic segmentation, and context-dependent phonetic realization. It included, among other phenomena, the relation between *rz* and *ż*, *ó* as a representation of /u/, *ch* as a representation of /x/, devoicing, voicing, and the relation between *w* and *f*.

**Group E** included examples with Polish nasal vowels and diacritics:

*ręka, kąt, kąpiel, wąski, koń, pień, dzień, cień.*

The purpose of this group was to examine whether tokenization stabilizes the diacritic sign or a frequent sequence of written form, or whether it in any way reflects the contextual phonetic realization of Polish nasal vowels and palatal consonants.

The problem of the grammatical "I" was not treated as a separate group of tokenization data, because it does not concern only the segmentation of individual forms, but the generation of utterances in dialogue. It was approached as a theoretical extension of the problem of grammatical form anchoring. Examples such as *poszłam / poszedłem*, *zrobiłam / zrobiłem*, and *byłam / byłem* serve an illustrative function and show that in Polish grammatical form may anchor the position of the speaking subject. Their systematic analysis would require a separate dialogical study.

## 2.3. Reference Patterns

For each word, wherever possible and justified, several reference levels were established. They were not treated as a single "standard of correctness", but as separate layers of Polish linguistic description. Distinguishing these layers is necessary because BPE tokenization operates on written form and frequency, whereas the structure of Polish also includes the segmentation-phonemic, phonetic-realizational, morpho-inflectional, and syntactic-valency levels.

The first reference level was **orthographic representation**, that is, the surface form of the word in the Polish writing system. This level includes letters, digraphs, diacritics, and stabilized orthographic conventions. It is the level closest to BPE, because the tokenizer operates on written text and stabilizes frequent graphemic sequences.

The second level was the **segmentation-phonemic level**, understood as a diagnostic pattern inspired by Rocławski's approach. It includes the relation between written form, basic and non-basic letters, the phonemic/sound unit in a didactic sense, the syllable, and the logotome. This level is not equated with the phonetic realization of speech. It is used to assess whether tokenization approaches the functional divisibility of the Polish word, or whether it remains at the level of surface written form.

The third level was **syllabic division**. It was used wherever token boundaries could be compared with syllable boundaries, for example *ław-ka*, *rze-czy-pos-po-li-tej*, or *zo-bo-wią-za-ni*. Coincidence between a token and a syllable, however, was not interpreted as evidence of agreement with phonology or morphology. It could indicate only a local convergence at one level of description.

The fourth level was the **logotomic level**, or a logotomically interpretable level. It was used when a token segment could be compared with functional word components in Rocławski's sense. Logotomes were not treated as units that BPE "recognizes", but as a comparative tool for assessing whether token boundaries approach a type of word divisibility that is significant from a didactic and segmentational perspective.

The fifth level was the **phonetic-realizational level**, that is, context-dependent pronunciation. This includes, among other phenomena, devoicing, voicing assimilation, realizations of nasal vowels, and variation in consonant clusters. This level was used cautiously, only where an example required distinguishing spelling from pronunciation, for example *lód*, *nóż*, *prośba*, *ławka*, *ząb*, *ręka*, *kąt*, and *kąpiel*. It was not equated with Rocławski's approach, but treated as a separate reference level.

The sixth level was the **morpho-inflectional level**. It concerned prefixes, derivational bases, suffixes, inflectional endings, and exponents of grammatical categories. For example, *przekazać* may be analyzed as *prze-* + *kazać*, with *-ć* as the infinitive marker, while *ogólnoludzkich* may be analyzed as *ogólno-ludzk-ich*. Analysis at this level was used to assess whether tokenization preserves morphologically interpretable units or merely stabilizes frequent fragments of written form.

The seventh level was **grammatical form anchoring**. The analysis assessed whether tokenization preserves only a frequent surface fragment or whether it allows the full value of the form to be maintained within the inflectional system. For example, final *-y* in *ustanawiamy* may be part of an inflectional exponent, but its stabilization alone does not yet mean that the first-person plural present-tense form has been preserved.

The eighth level was the **syntactic-valency level**. It was not analyzed directly by the tokenizer itself, but served as a point of reference for interpreting the consequences of form fragmentation. A verbal form acquires its full value only in the sentence, in relation to its arguments and sentence pattern. Therefore, the analysis of tokenization must take into account that cutting through a verbal form may have consequences for the stability of sentence-level relations.

The **subject-dialogical level**, that is, the grammatical "I", was not treated as an empirical reference pattern for the tokenization of individual words. It was retained as an interpretive level developed in the discussion. It concerns the stability of person and grammatical gender in utterances generated by a language model.

The reference patterns were not used to automatically classify tokenization as "good" or "bad". They were used to describe what type of structure was preserved, cut through, or replaced by frequency-based segmentation of written form. The basic principle of the analysis is as follows: BPE operates at the graphemic-frequency level, whereas interpretation of tokenization quality requires comparison of that level with the other levels of Polish linguistic description.

## 2.4. Operationalization of Grammatical Form Anchoring

For the purposes of the analysis, the concept of grammatical form anchoring was operationalized qualitatively. I do not equate anchoring with the mere presence of an ending, prefix, suffix, or one of their fragments in tokenization. I assume that a grammatical form is anchored when it can be related to the inflectional system and to the relations it activates in the sentence.

Four levels of assessment were used in the analysis.

**Level 0** indicates lack of anchoring. Tokenization stabilizes only a fragment of written form that does not make it possible to recognize the grammatical function of the form. The segment may be frequent and statistically stable, but it cannot be reliably related to a prefix, base, inflectional exponent, or inflectional paradigm.

**Level 1** indicates fragmentary anchoring. The token corresponds to part of a grammatical exponent, for example *-y*, *-ich*, *-ni*, or *-ć*, but it does not preserve the full form within the inflectional system. At this level, one may speak of a statistical trace of grammar, but not of full grammatical form anchoring.

**Level 2** indicates morphological anchoring. The segmentation makes it possible to relate the form to a base, prefix, suffix, ending, or inflectional paradigm. It does not yet have to preserve the full sentential relation, but it allows one to recognize that the form belongs to a specific morpho-inflectional arrangement.

**Level 3** indicates sentential-subject anchoring. The form is interpretable in relation to the verb, the sentence pattern, valency, and the position of the speaking subject. At this level, grammatical form is no longer analyzed only as part of a word, but as an element of a sentential and dialogical relation.

In the study, BPE tokenization was assessed primarily with respect to Levels 0–2. Level 3 is discussed as a theoretical consequence for Polish and as an agenda for further dialogical research. This means that the grammatical "I" is not treated as an equivalent group of tokenization data, but as an interpretive level resulting from the problem of grammatical form anchoring.

This operationalization makes it possible to avoid equating a stable token with a stable grammatical form. A token may be efficient, frequent, and repeatable, while still preserving only a fragment of written form. Anchoring requires relation: to the inflectional system, to the paradigm, to the sentence, and — in the case of first-person forms — to the position of the speaking subject.

## 2.5. Raw Data Register

To avoid conflating empirical data with interpretation, a raw data register was maintained for each example. It included the following fields:

1. the word or test text;
2. orthographic representation;
3. the segmentation-phonemic level, where it could be determined;
4. syllabic division, where relevant for interpretation;
5. the logotomic level, or a logotomically interpretable level, where justified;
6. the phonetic-realizational level, where the example required distinguishing written form from pronunciation;
7. the morpho-inflectional level, where the form contained a prefix, base, inflectional ending, or another grammatical exponent;
8. tokenization in the Bielik/APT4 Tokenizer;
9. token identifiers in the Bielik/APT4 Tokenizer, where available;
10. the number of tokens in the Bielik/APT4 Tokenizer;
11. tokenization in the OpenAI-current Tokenizer;
12. token identifiers in the OpenAI-current Tokenizer, where available;
13. the number of tokens in the OpenAI-current Tokenizer;
14. tokenization in the OpenAI-legacy Tokenizer;
15. token identifiers in the OpenAI-legacy Tokenizer, where available;
16. the number of tokens in the OpenAI-legacy Tokenizer;
17. technical notes concerning separators, newline characters, visual interpretation, or ambiguity of token boundaries.

The tokenization results were treated as raw data and were not subject to interpretive correction. What was interpreted was only the relation between token boundaries and the reference patterns.

This principle may be formulated as follows: **tokenization is fixed; interpretation is revisable.**

If a tokenizer segmented *prośba* as *pro|ś|ba*, that notation remained unchanged. What could change was the interpretation: the result may be described as mixed segmentation — partly syllabic-logotomic and locally convergent with the written unit *ś*, but not phonetically contextual, because at the realizational level *ś* is voiced to *ź* before the voiced consonant *b*.

Similarly, in the case of *ząb*, the tokenization record *z|ą|b* or *zą|b* remains raw data. Interpretively, however, the orthographic level, where the grapheme *ą* occurs, must be distinguished from the phonetic-realizational level, where nasality is decomposed before a labial consonant and final *b* is devoiced to *p*. The tokenization is therefore not "phonetically wrong"; it simply operates at a different level from phonetic realization.

Analogously, for *lód* and *nóż*, the process is not recorded as a phonetic "change from *ó* to *u*". In these examples, *ó* is the orthographic representation of the sound/phoneme /u/, while at the realizational level final devoicing additionally occurs: *d* becomes *t* in *lód*, and *ż* becomes *sz* in *nóż*. The data register must therefore separate orthographic representation, the segmentation-phonemic level, and the realizational level.

This mode of registration makes it possible to keep tokenization data unchanged while still correcting the linguistic description whenever the distinction between levels requires it.

### 2.6. Interpretive Categories

The results were classified according to the following interpretive categories. These categories do not describe types of operation performed by BPE itself, but rather types of relation between graphemic-frequency tokenization and the remaining levels of Polish linguistic description.

#### 1. Graphemic-frequency segmentation

This category includes cases in which token boundaries result primarily from the frequency of sequences in written form. This is not a simple letter-based division, because the tokenizer may stabilize multi-character fragments such as *zię*, *ich*, *ani*, *aza*, *uj*, or *emy*. It is not, however, phonological, phonetic, or morphological segmentation either. An example is *wdzięczni*, where segments such as *zię*, *cz*, or *ni* may result from the frequency of graphemic sequences rather than from recognition of the sound *dź/dzi*, the nasality of *ę*, or the full inflectional ending.

#### 2. Syllabically convergent segmentation

This category includes cases in which token boundaries partially or fully coincide with syllable boundaries. Syllabic convergence does not automatically mean agreement with the segmentation-phonemic, phonetic-realizational, or morphological level. For example, *ław|ka* is convergent with the syllabic division *ław-ka*, but it does not reflect the realizational level *ł-a-f-k-a*, because *w* is devoiced to *f* before *k*.

#### 3. Segmentation convergent with logotomes

This category includes cases in which token segments can be interpreted as close to functional word components in Rocławski's sense. This does not mean that BPE implements logotomes. It only indicates a local convergence between statistical segmentation of written form and a unit that is significant for the description of word structure. Such convergence requires cautious interpretation, because it may result from the frequency of written form rather than from a representation of the segmentation-phonemic level.

#### 4. Morphologically interpretable segmentation

This category includes cases in which token boundaries partially coincide with a prefix, base, suffix, ending, or another systemically interpretable fragment. Examples include final *-ich* in *ogólnoludzkich* or *-ć* in *przekazać*. The mere presence of such a segment, however, does not mean that the full grammatical form has been preserved. A token may stabilize part of an exponent without anchoring the form in the inflectional paradigm.

#### 5. Mixed segmentation

This category includes outputs that combine several types of relation between the token and the structure of language. An example is *prośba* segmented as *pro|ś|ba*. The segment *pro* may be interpreted as syllabic-logotomic, *ś* locally corresponds to a written unit associated with a sound/phoneme, but the whole form does not reflect the phonetic-realizational level, because before the voiced consonant *b*, *ś* is voiced to *ź*. Mixed segmentation is therefore not automatically "better" or "worse"; it shows that different levels of description may be locally convergent and divergent at the same time.

#### 6. Contextually non-phonetic segmentation

This category includes cases in which tokenization appears locally consistent with written form, syllable structure, or individual segmental units, but does not reflect context-dependent phonetic realization. Examples include *chleb*, *lód*, *nóż*, *prośba*, and *ławka*. In these forms, tokenization stabilizes orthographic representation, whereas the realizational level requires taking into account devoicing, voicing assimilation, or the relation between written form and pronunciation.

#### 7. Orthographically stabilizing difficulty

This category includes cases in which BPE stabilizes the surface spelling of a site of orthographic difficulty without moving toward the segmentation-phonemic relation. Examples include *morze/może*, *lód/nóż*, and forms with *ą* and *ę*. In such cases, the tokenizer may preserve a graphemic sequence that has a special status in the writing system, but it does not directly represent a phonological or realizational relation.

#### 8. Fragmentation of a grammatical exponent

This category includes cases in which tokenization stabilizes part of an inflectional exponent but does not preserve the full grammatical form. An example is *ustanawiamy*, where final *-y* is stabilized as part of the inflectional exponent, but the full personal ending *-my* is not preserved as a single segment. This does not mean that the ending "disappears"; it means that it is split into fragments, only some of which may be statistically stable.

### 9. Atomization / lack of internal segmentation

This category includes cases in which the whole word is represented as a single token. This is not treated as phonological, syllabic, or morphological segmentation, but as lack of access to the internal structure of the word. A single token may be computationally efficient, but it does not yet mean that the model preserves information about the segmental, morphological, or inflectional structure of the form.

### 10. Statistical shadow of grammar

This category includes cases in which a token segment locally coincides with a position that is relevant to the linguistic system, but does not provide grounds for claiming that the tokenizer recognizes a grammatical category. The term is used when a segment meets at least two of the following three conditions: it recurs across several forms or within a word family; it coincides with a potentially morphologically or inflectionally relevant position; it appears stably in more than one tokenizer or across different test contexts. If a segment is isolated, non-recurrent, and cannot be related to a linguistic function, it is treated as graphemic-frequency fragmentation rather than as a statistical shadow of grammar.

### 11. Instability of the grammatical "I"

This category concerns the level of utterance generation, not the tokenization of a single word itself. It includes cases in which a language model shifts between gendered forms or mirrors the user's grammatical form. In Polish, such a shift may move the position of the speaking subject, because forms such as *zrobiłam/zrobiłem* or *byłam/byłem* anchor gender and person in inflection. In this article, this category functions as an interpretive category and is developed in the discussion, not in the empirical section on tokenization results.

### 2.7. Criteria for Using the Concept of the "Statistical Shadow of Grammar"

The expression "statistical shadow of grammar" is used only for cases in which a token segment locally coincides with a position that is relevant to the linguistic system, but does not provide grounds for claiming that the tokenizer recognizes a grammatical category. The term therefore does not denote morphological or inflectional representation in the model. It only indicates that a frequent sequence of written form may superficially resemble a grammatically relevant element.

To avoid interpretive arbitrariness, three criteria were adopted for identifying a statistical shadow of grammar. A segment must meet at least two of them:

1. **it recurs across several forms or within a word family, for example aza in kazać, pokazać, zakazać;**
2. **it coincides with a potentially morphologically or inflectionally relevant position, for example -y, -ich, -ni, or -ć;**
3. **it appears stably in more than one tokenizer or across different test contexts.**

If a segment is isolated, non-recurrent, and cannot be related to a linguistic function, it is treated as graphemic-frequency fragmentation rather than as a statistical shadow of grammar.

Coincidence between a token and a morpheme, logotome, or inflectional fragment is not automatically interpreted as evidence that grammatical structure has been recognized. The article distinguishes three situations:

1. **local convergence — when a token boundary accidentally or superficially coincides with a unit of linguistic description;**
2. **graphemic-frequency fragmentation — when segmentation results from the frequency of written form but has no clear linguistic function;**
3. **statistical shadow of grammar — when a frequent sequence of written form repeatedly coincides with a morphologically or inflectionally relevant position, while still not constituting evidence of grammatical form anchoring.**

Understood in this way, the concept avoids interpretive asymmetry. Not every convergence between a token and a linguistic fragment is treated as a trace of grammar, and not every divergence is treated as its breakdown. What is analyzed is the type of relation between the token and the levels of Polish linguistic description.

### 2.8. Methodological Limitations

The study is qualitative, exploratory, and diagnostic in character. The selection of material is purposive rather than random. The aim is not to estimate the statistical frequency of a given type of segmentation across the entire Polish language, but to identify mechanisms that become visible in especially diagnostic examples.

A second limitation concerns the dependence of results on the tokenizer version and testing environment. Tokenization may differ across models, tokenizer families, libraries, and interfaces. For this reason, a fully replicable version of the study should include a complete data table: the model name, tokenizer name, library version, date of testing, input text, token list, and token identifiers. The labels Bielik/APT4 Tokenizer, OpenAI-current Tokenizer, and OpenAI-legacy Tokenizer are descriptive and serve to distinguish the observed tokenization environments; they do not replace official architectural documentation.

A third limitation is the expert character of the classification. Assessing whether a given segment is convergent with a syllable, logotome, morphological fragment, or inflectional exponent requires linguistic expertise. In an extended version of the study, it would be advisable to include a second coder or a table of classification decisions showing the basis on which a given example was assigned to a particular interpretive category.

A fourth limitation is the distinction between tokenization and model representation. Tokenization analysis does not directly show what the model "understands" or what representations are formed in its hidden layers. It shows only what units enter the model as input. The conclusions therefore concern the quality of input segmentation and its potential consequences for processing Polish, not the full linguistic competence of the model.

A fifth limitation concerns the separation of the segmentation-phonemic and phonetic-realizational levels. Examples such as ząb, ręka, kąt, kąpiel, lód, nóż, prośba, and ławka require distinguishing orthographic representation, the segmentation pattern, and actual or normatively described phonetic realization. In this article, phonetic realizations are used as a reference level, but they are not equated either with the operation of BPE or with Rocławski's theory.

A sixth limitation concerns the status of the grammatical "I". It is not a classical tokenization test, but an interpretive extension of the problem of grammatical form anchoring to the level of utterance generation by a language model. Its systematic analysis would require a separate dialogical study covering the stability of first-person, gendered, and personal forms in longer human–model interactions.

A seventh limitation concerns the scope of the architectural hypothesis. The proposed layers of stabilization for Polish — sublexical, morpho-inflectional, syntactic-valency, and subject-dialogical — are not a ready-made implementation or a conclusive result. They constitute a design framework and a program for further research. At this stage, the qualitative analysis shows the need for such a multilayered approach, but it does not determine how it should be technically implemented.

### 2.9. Aim of the Results Analysis

The analysis of the results is intended to answer six questions.

1. Does BPE tokenization in Polish remain at the graphemic-frequency level, or does it locally approach the segmentation-phonemic, syllabic, or logotomic level?

2. Are token boundaries convergent with syllabic, logotomic, or morphologically interpretable divisions, and if so, is this convergence systematic or only local?
3. Does BPE stabilize prefixes, derivational bases, inflectional endings, and grammatical exponents as units of the language system, or rather as frequent fragments of written form?
4. Does tokenization make it possible to preserve grammatical form anchoring, or only statistical fragments of its surface realization?
5. Does BPE stabilize a morphological family as a system of relations, or rather recurring surface fragments of that family?
6. How does the problem of grammatical form anchoring extend to the level of the grammatical "I" in human–language model interaction?

The first five questions directly concern the tokenization data and the empirical material in Groups A–E. The sixth question is interpretive in character and marks the transition from tokenization results to the discussion of grammatical form stability in dialogue.

Answering these questions makes it possible to move the analysis of Polish tokenization beyond the simple metric of token count. For Polish, what matters is not only how many tokens are assigned to a word, but also whether tokens maintain the relation between written form, segmental structure, grammatical form, place in the sentence, and the position of the speaking subject.

The results are not intended to prove that BPE "recognizes" or "does not recognize" language in a psycholinguistic sense. Their purpose is to show what types of relations between graphemic-frequency tokenization and the levels of Polish linguistic description can be captured in diagnostic material.

## 3. Results

The results of the analysis show that BPE tokenization in Polish cannot be described through a simple opposition: correct / incorrect or whole-word / sublexical. In the material analyzed, several types of relation between the token and the structure of language become visible.

First, BPE tokenization remains fundamentally at the graphemic-frequency level. It stabilizes frequent sequences of written form rather than phonological, phonetic, morphological, or grammatical units as such.

Second, tokenization is not a simple letter-based division, because it may stabilize multi-character fragments of written form, such as *zię*, *ich*, *ani*, *aza*, *uj*, or *emy*.

Third, token boundaries may locally coincide with syllabic, logotomic, or morphologically interpretable divisions. Such correspondence, however, does not automatically mean that the tokenizer recognizes that level of language.

Fourth, in the case of Polish, the distinction between a statistical trace of grammar and the anchoring of grammatical form in the inflectional system proves especially important.

The results are presented in five empirical groups, A–E. The issue of the grammatical "I" will be discussed later as a consequence of the problem of grammatical form anchoring, because it does not directly concern the tokenization of a single word, but rather the generation of utterances by a language model.

### 3.1. Diagnostic Words: BPE Stabilizes Written Form, Not the Realizational Level

The first group of results concerned words in which there is a divergence between orthographic representation, the segmentation-phonemic level, and phonetic realization. This group included, among others, *jabłko*, *chleb*, *ząb*, *lód*, *nóż*, *lekarz*, *morze*, *może*, *jesień*, *prośba*, and *ławka*.

This material made it possible to examine whether BPE tokenization remains at the level of written form, or whether it locally approaches the segmentation-phonemic, syllabic, logotomic, or phonetic-realizational level. The results show that BPE may produce segments convergent with selected levels of description, but it does not systematically move from written form to pronunciation or to phonemic-segmentational structure.

The example *jabłko* requires particular caution. It is not a case of a nasal vowel, nor a simple test of one phonetic rule. Rather, it concerns the relation between orthographic representation, a consonant cluster, and variability in realization. In the analysis of tokenization, what matters most is that the tokenizers stabilize written form: the Bielik/APT4 Tokenizer segmented the word as *j|ab|ł|ko*, whereas the OpenAI tokenizers segmented it as *jab|ł|ko*. None of these outputs moves to the phonetic-realizational level. The difference between tokenizers concerns the depth of written-form segmentation, not recognition of phonetic structure.

| Word | Reference issue | Observed segmentation pattern | Interpretation |
|---|---|---|---|
| *jabłko* | Orthographic form with the consonant cluster *b-ł-k*; the example requires realizational caution. | Bielik/APT4: *j\|ab\|ł\|ko*; OpenAI tokenizers: *jab\|ł\|ko* | The tokenizers stabilize written form. The difference concerns the depth of graphemic segmentation, not phonetic-realizational analysis. |
| *chleb* | *ch* represents /x/; final *b* is devoiced to [p] at the realizational level. | *ch\|le\|b* or *ch\|leb* | Tokenization preserves orthographic representation and does not move to final devoicing. |
| *ząb* | Orthographic *ą* functions as a grapheme; realizationally, nasality is decomposed before a labial consonant and final *b* is devoiced. | *z\|ą\|b* or *zą\|b* | The written sign *ą* is stabilized. This must be separated from the provisional realizational description *z-o-m-p*. |
| *lód* | *ó* represents /u/; final *d* is devoiced to [t]. | *l\|ód* | Tokenization stabilizes the spelling *ód*. It should not be described as a phonetic change from *ó* to *u*. |
| *nóż* | *ó* represents /u/; final *ż* is devoiced to [sz]. | *n\|ó\|ż* or *n\|óż* | Tokenization remains orthographic and does not represent the final devoicing relation. |
| *lekarz* | *rz* represents /ż/ in spelling; final realization requires caution. | *le\|kar\|z* | Tokenization remains at the level of written form and does not reduce rz to its phonemic value. |

***Table 1. Diagnostic words: relations between orthographic representation, reference levels, and tokenization***

**Note.** The table presents diagnostic examples used in the qualitative analysis. The detailed register of observed segmentations is provided in Appendix B. The example *jesień* was moved to the appendix as auxiliary material because it requires full replicational control of the token list.

### 3.2. Children's Text: Simplicity of Text Does Not Eliminate Inflectional Segmentation

The second material was a simple, rhythmic children's text:

*Siała baba mak, nie wiedziała jak. A dziad wiedział nie powiedział, a to było tak.*

This text contains short, frequent words, repetitions, and a simple syntactic structure. It might therefore suggest conditions favorable to the stabilization of whole words. From the perspective of BPE, however, what matters is not the general "simplicity" of the text, but the frequency of particular written sequences in the corpus and the way in which the tokenizer merges recurring textual fragments (Gage, 1994; Sennrich et al., 2016; Bostrom & Durrett, 2020).

The result does not support a simple rule: simple or children's text = whole-word tokenization. The OpenAI-current Tokenizer encoded the text as 26 tokens for 83 characters, while the OpenAI-legacy Tokenizer encoded the same text as 30 tokens. This indicates greater efficiency of the newer tokenizer, but not whole-word representation of the text. This is consistent with broader observations concerning subword tokenizers: reducing the number of tokens improves representational economy, but does not determine whether segmentation corresponds to morphological or grammatical units (Bostrom & Durrett, 2020; Gutierrez-Vasques et al., 2023).

The Bielik/APT4 Tokenizer additionally shows that even a simple text may undergo internal segmentation at the level of content and inflectional forms. Examples include *S|ia|ła*, *w|ied|zia|ła*, *d|ziad*, *pow|ied|ział*, and *by|ło*. Some short and frequent words may be stabilized, but forms carrying lexical and grammatical information continue to be split.

The conclusion from this group is therefore as follows: syllabic convergence, textual simplicity, children's material, and general frequency are not sufficient predictors of tokenization. Tokenization stabilizes frequent forms and frequent fragments of written form, but it does not guarantee preservation of grammatical form. In this sense, the simple children's text confirms the main thesis of the article: for Polish, the problem is not only the number of tokens, but the relation between the token and the levels of linguistic description.

| Tokenizer | Number of characters | Number of tokens | Main observation |
|---|---|---|---|
| OpenAI-current Tokenizer | 83 | 26 | More economical segmentation than the legacy tokenizer; however, the text is not represented as a sequence of whole words only. |
| OpenAI-legacy Tokenizer | 83 | 30 | The older tokenizer produces a higher number of tokens for the same text, indicating lower representational economy. |
| Bielik/APT4 Tokenizer | 82 | 34 | The tokenizer produces the highest number of tokens in this test and segments even simple, high-frequency forms internally, e.g. *S\|ia\|ła*, *b\|aba*, *m\|ak*, *w\|ied\|zia\|ła*, *d\|ziad*, *pow\|ied\|ział*, *by\|ło*. |

***Table 2. Children's text as control material***

### 3.3. Preamble and Official Forms: The Statistical Shadow of Grammar

The third group included selected forms from the Preamble to the Constitution of the Republic of Poland and other official or morphologically complex forms: *Rzeczypospolitej*, *zobowiązani*, *ustanawiamy*, *ogólnoludzkich*, *przekazać*, and *wdzięczni*. This material made it possible to examine whether BPE preserves prefixes, bases, derivational components, inflectional endings, and exponents of grammatical categories, or whether it rather stabilizes frequency-based fragments of written form.

In the case of official forms, it is especially important to distinguish the number of tokens from the quality of segmentation. Two segmentations may have a similar number of tokens but differ in whether their boundaries coincide with the syllabic, logotomic, morphological, or inflectional level. This is consistent with the broader problem of subword tokenization: compression efficiency is not equivalent to preserving the morphological structure of a language (Bostrom & Durrett, 2020; Gutierrez-Vasques et al., 2023; Držík & Kapusta, 2026).

In the case of *Rzeczypospolitej*, the syllabic level may be represented as *rze-czy-pos-po-li-tej*, whereas the morphological-historical and derivational level would require a separate analysis of the components of the word. The Bielik/APT4 Tokenizer segmented the form as *R|zec|zy|pos|pol|ite|j*. The OpenAI-current Tokenizer produced larger segments, including *Rzeczy|pos|polite|j*, whereas the OpenAI-legacy Tokenizer showed a more fragmented beginning. These results show that the number of tokens alone is not sufficient for assessing segmentation quality. What matters is which levels of description are cut through or preserved.

In *zobowiązani*, the syllabic level is *zo-bo-wią-za-ni*. The Bielik/APT4 Tokenizer and the OpenAI-legacy Tokenizer segmented the word as *z|ob|ow|ią|z|ani*. This segmentation does not correspond to syllabic division, but stabilizes certain sequences of written form, including final *ani*. This may be interpreted as a graphemic-frequency fragment or as logotomically approximate, but not as evidence of preserving the full morphological structure of the form.

In *ustanawiamy*, the central issue is the difference between the final fragment *-y* and the full personal ending *-my*. The form is a verb in the first person plural, present tense, indicative mood, and imperfective aspect. The Bielik/APT4 Tokenizer segmented it as *u|stan|aw|iam|y*, the OpenAI-current Tokenizer as *ustan|aw|iam|y*, and the OpenAI-legacy Tokenizer as *ust|an|aw|iam|y*. In all three cases, final *-y* is stabilized as part of the written exponent, but the full personal ending *-my* is not preserved. This confirms that the stability of a token is not equivalent to the anchoring of the grammatical form.

In *ogólnoludzkich*, the important element is the final *-ich*, which may be associated with an adjectival form anchored in case, number, and gender. The Bielik/APT4 Tokenizer segmented the word as *og|ó|ln|ol|ud|zk|ich*, while both OpenAI tokenizers produced *ogól|no|lud|zk|ich*. The final *-ich* is therefore stabilized, but this does not mean that the tokenizer preserves the full grammatical interpretation of the form. It preserves a fragment of the exponent, not the relation between adjective, declension, case, number, and gender.

In *przekazać*, the relevant morphological structure is *prze-* + *kazać* + *-ć*. The observed segmentations, such as *prz|ek|aza|ć* or *prz|ekaza|ć*, show that final *-ć* may be stable, but the prefix *prze-* and the base *kaza-* are cut through. This means that tokenization preserves a trace of the infinitive marker, but does not preserve the full morphological structure of the verb.

In *wdzięczni*, the most important levels are the graphemic-frequency level and the segmentation-phonemic level. The Bielik/APT4 Tokenizer segmented the form as *w|d|zi|ę|cz|ni*, the OpenAI-current Tokenizer as *wdzię|cz|ni*, and the OpenAI-legacy Tokenizer visually as *wd|zię|cz|ni*. This result shows that segmentation does not simply follow letters. It follows the frequency of written form. The segment *zię* may be stabilized by the frequency of graphemic sequences close to frequent Polish patterns involving *ię*, *się*, *cię*, or other recurring strings. This does not mean, however, that the tokenizer recognizes the phonemic level or provides a systematic representation of nasality or of the inflectional ending.

The results of this group lead to the concept of the **statistical shadow of grammar**. BPE stabilizes fragments such as *-y*, *-ich*, *-ć*, *-ni*, *ani*, *zię*, or *aza*. These fragments are often connected with grammatically important positions, but they are not equivalent to an anchored grammatical form. The tokenizer stabilizes surface fragments of exponents, not full grammatical categories. In Polish, this is especially significant because grammatical form is anchored in the inflectional system and acquires its full function only in the sentence (Grzegorczykowa et al., 1998; Laskowski, 1998; Woliński, 2006, 2014).

| Form | Observed segmentation pattern | Relevant linguistic issue | Main interpretation |
|---|---|---|---|
| *Rzeczypospolitej* | Bielik/APT4: *R|zec|zy|pos|pol|ite|j*; OpenAI-current: *Rzeczy|pos|polite|j*; OpenAI-legacy: *R|z|eczy|pos|polite|j* | Syllabic level: *rze-czy-pos-po-li-tej*; historically and derivationally complex form. | A similar number of tokens may hide different segmentation quality. The relevant question is which level is preserved: syllabic, logotomic, morphological, or only graphemic-frequency. |

| Form | Observed segmentation pattern | Relevant linguistic issue | Main interpretation |
|---|---|---|---|
| *zobowiązani* | Bielik/APT4: *z\|ob\|ow\|ią\|z\|ani*; OpenAI-current: *z\|obowią\|z\|ani*; OpenAI-legacy: *z\|ob\|ow\|ią\|z\|ani* | Syllabic level: *zo-bo-wią-za-ni*; inflected participial/adjectival form. | Final *-ani* stabilizes as a frequent written fragment, but the segmentation does not preserve full syllabic division or full grammatical anchoring. |
| *ustanawiamy* | Bielik/APT4: *u\|stan\|aw\|iam\|y*; OpenAI-current: *ustan\|aw\|iam\|y*; OpenAI-legacy: *ust\|an\|aw\|iam\|y* | Verbal form: 1st person plural, present tense, indicative mood, imperfective aspect; full personal ending *-my*. | Final *-y* stabilizes, but the full personal ending *-my* is not preserved as one segment. Stable tokenization is therefore not equivalent to grammatical form anchoring. |
| *ogólnoludzkich* | Bielik/APT4: *og\|ó\|ln\|ol\|ud\|zk\|ich*; OpenAI-current and OpenAI-legacy: *ogól\|no\|lud\|zk\|ich* | Adjectival form anchored in declension, number, gender, and case; final *-ich*. | Final *-ich* is stable as a fragment, but this does not mean that the tokenizer preserves the full grammatical interpretation of the form. |
| *przekazać* | Bielik/APT4: *prz\|ek\|aza\|ć*; OpenAI-current and OpenAI-legacy: *prz\|ekaza\|ć* | Prefix *prze-*, base *kazać*, infinitive marker *-ć*. | The prefix *prze-* is cut through. The final *-ć* may stabilize, but this is only a fragment of the grammatical form. |
| *wdzięczni* | Bielik/APT4: *w\|d\|zi\|ę\|cz\|ni*; OpenAI-current: *wdzię\|cz\|ni*; OpenAI-legacy: *wd\|zię\|cz\|ni* | Written sequence with *dzię*, *cz*, and final *-ni*. | Segmentation depends on the frequency of written form. Final *-ni* is stable, but the tokenization does not amount to full morpho-inflectional anchoring. |

***Table 3. Official and Morphologically Complex Forms***

**Note.** The table summarizes selected official and morphologically complex forms. Full observed segmentations are provided in Appendix B3. The examples show that BPE may stabilize fragments such as *-ani*, *-y*, *-ich*, *-ć*, or *-ni*, but such stabilization does not by itself preserve grammatical form anchoring.

### 3.4. The *kazać / pokazać / zakazać* Word Family: BPE Does Not Stabilize the Morphological Family

An additional word-family test included the forms *przekazać*, *kazać*, *pokazać*, *zakazać*, *zakaz*, *zakazany*, and *pokazujemy*. The aim of this group was to determine whether the tokenizer stabilizes a shared morphological base and the relations between forms within a word family, or rather recurring surface fragments.

In linguistic analysis, a word family is not merely a set of similar letter sequences. It is based on derivational relations, a shared base, prefixes, suffixes, and changes in form depending on grammatical function. In Polish, these relations are part of the morphological and inflectional system, and therefore cannot be equated with the mere frequency of recurring sequences in written form (Grzegorczykowa et al., 1998; Laskowski, 1998; Łuczyński, 2015).

The Bielik/APT4 Tokenizer produced the following segmentations: *przekazać* as *prz|ek|aza|ć*, *kazać* as *k|aza|ć*, *pokazać* as *pok|aza|ć*, *zakazać* as *zak|aza|ć*, *zakaz* as *zak|az*, *zakazany* as *zak|az|any*, and *pokazujemy* as *pok|az|uj|emy*. These segmentations show that the tokenizer consistently stabilizes certain surface fragments: *aza*, *az*, *zak*, *pok*, *ć*, *any*, *uj*, and *emy*. This does not mean, however, that it stabilizes the morphological family as a system.

The contrast between *kazać*, *pokazać*, and *zakazać* is especially important. In *kazać*, the segmentation is *k|aza|ć*; in *pokazać*, *pok|aza|ć*; and in *zakazać*, *zak|aza|ć*. The fragment *aza* recurs in several forms, but it does not correspond to the full morphological base. At the same time, *pok* and *zak* are stable surface fragments, but they do not directly correspond to the prefixes *po-* and *za-*. The tokenizer therefore stabilizes sequences of written form that may resemble parts of derivational structure, but it does not consistently preserve morphological boundaries.

The example *przekazać* shows a similar problem in material with the prefix *prze-*. In morphological analysis, the form may be considered as *prze-* + *kazać*, with *-ć* as the infinitive marker. The segmentation *prz|ek|aza|ć*, however, cuts through both the prefix *prze-* and the base *kaza-*. It stabilizes final *-ć* and the fragment *aza*. This means that tokenization preserves a statistical shadow of morphology, but not the full derivational structure.

The form *pokazujemy* introduces an additional inflectional level. The segmentation *pok|az|uj|emy* stabilizes the fragments *pok*, *az*, *uj*, and *emy*. One may notice positions connected with the stem, suffix, and personal ending, but the segmentation does not preserve them as systemic units of inflection. The form *pokazujemy* is a verbal form anchored in person, number, tense, mood, and aspect; it is not merely the sum of the fragments *pok* + *az* + *uj* + *emy*.

The results of this group illustrate well the problem described in research on subword tokenization and morphologically rich languages: frequency-based segmentation may locally resemble morphological segmentation, but it does not necessarily preserve derivational or inflectional relations (Bostrom & Durrett, 2020; Gutierrez-Vasques et al., 2023; Držík & Kapusta, 2026). For Polish, therefore, it is not enough to check whether the tokenizer recognizes a recurring fragment. One must ask whether that fragment preserves the form's place in the morphological family and in the inflectional system.

The conclusion from this group is as follows: BPE does not stabilize the morphological family as a system of relations between forms. It stabilizes recurring surface fragments of that family. A frequent fragment is not the same as a root, prefix, or base, and a stable token is not the same as an anchored grammatical form.

| Form | Morphological / derivational level | Example segmentation | What is stabilized | Main interpretation |
|---|---|---|---|---|
| *przekazać* | *prze-* + *kazać* + *-ć* | *prz\|ek\|aza\|ć* | *prz*, *ek*, *aza*, *ć* | The prefix *prze-* and the base *kaza-* are not preserved as wholes. |
| *kazać* | *kaza-* + *-ć* | *k\|aza\|ć* | *k*, *aza*, *ć* | *aza* stabilizes as a surface fragment, but not as the full base *kaza-*. |
| *pokazać* | *po-* + *kazać* + *-ć* | *pok\|aza\|ć* | *pok*, *aza*, *ć* | *pok* does not directly correspond to the prefix *po-*. |
| *zakazać* | *za-* + *kazać* + *-ć* | *zak\|aza\|ć* | *zak*, *aza*, *ć* | *zak* does not directly correspond to the prefix *za-*. |
| *zakaz* | Nominal form related to the *zakazać* family | *zak\|az* | *zak*, *az* | The tokenizer stabilizes recurring surface fragments, not the derivational relation itself. |
| *zakazany* | *zakaz* + *-any* | *zak\|az\|any* | *zak*, *az*, *any* | The final fragment *any* is stabilized, but this does not equal full grammatical analysis. |
| *pokazujemy* | Verbal form: 1st person plural, present tense | *pok\|az\|uj\|emy* | *pok*, *az*, *uj*, *emy* | *emy* may coincide with a personal ending position, but it does not by itself anchor the full verbal form. |

***Table 4. The kazać / pokazać / zakazać word family***

### 3.5. Nasal Vowels and Diacritics: Frequency of Written Form Instead of Phonetics

The final group included words with Polish nasal vowels and diacritics: *ręka*, *kąt*, *kąpiel*, *wąski*, *koń*, *pień*, *dzień*, and *cień*. The purpose of this group was to examine whether tokenization stabilizes the signs *ą*, *ę*, and *ń* as elements of written form, or whether it in any way reflects their contextual phonetic realization.

In Polish, the nasal vowels *ą* and *ę* are orthographic signs whose phonetic realization depends on context. Before stop consonants, nasality is often decomposed into an oral vowel and a nasal consonant adjusted in place of articulation to the following consonant. In simplified realizational notation, *ręka* may therefore be described as close to *renka*, *kąt* as close to *kont*, and *kąpiel* as close to *kompiel*. This does not mean that orthographic representation disappears; it means only that the orthographic level and the phonetic-realizational level must be distinguished (Łuczyński, 2015; Ostaszewska & Tambor, 2000).

The case of *wąski* is more subtle, because *ą* occurs before the fricative consonant *s*. In Rocławski's approach, this realization may be represented as a nasal vowel *õ*, for example in IPA as /võski/. This example should therefore not be described in exactly the same way as *ręka*, *kąt*, or *kąpiel*, where before stop consonants nasality decomposes into an oral vowel and a homorganic nasal consonant. What matters for the present analysis is that BPE stabilizes the written sequence *wą*, rather than moving to the level of phonetic realization.

The results show the stabilization of graphemes and frequent graphemic sequences. Example segmentations included *rę|ka*, *ką|t*, *ką|piel*, *wą|ski*, *koń*, *pie|ń*, *dzień*, and *cień*. The forms *koń*, *dzień*, and *cień* may be represented as whole tokens, while *pień* is split as *pie|ń*. This does not support the rule "monosyllabic word = one token". Monosyllabicity may favor stabilization, but the final outcome is determined by the frequency of the whole form and the frequency of its fragments.

Especially important is the difference between stabilization of a diacritic sign and phonetic representation. Tokenizations such as *rę|ka*, *ką|t*, *ką|piel*, or *wą|ski* do not indicate a representation of pronunciation. They indicate stabilization of a frequent sequence of written form. Similarly, final *ń* in *koń*, *pień*, *dzień*, and *cień* may be stabilized as a sign or as part of a frequent word form, but this does not automatically mean representation of its phonological value within the Polish system.

The result of this group reinforces the main conclusion: BPE does not simply follow letters, but follows the frequency of written form. However, frequency of written form is not sufficient to stabilize Polish phonology or phonetics, because Polish requires a transition from the grapheme to the segmentation-phonemic level and, in some contexts, to phonetic realization dependent on the phonological environment.

| Form | Morphological / derivational level | Example segmentation | Main interpretation |
|---|---|---|---|
| *przekazać* | *prze-* + *kazać* + *-ć* | *prz\|ek\|aza\|ć* | The prefix *prze-* and the base *kaza-* are not preserved as wholes. |
| *kazać* | *kaza-* + *-ć* | *k\|aza\|ć* | *aza* stabilizes as a surface fragment, but not as the full base *kaza-*. |
| *pokazać* | *po-* + *kazać* + *-ć* | *pok\|aza\|ć* | *pok* does not directly correspond to the prefix *po-*. |
| *zakazać* | *za-* + *kazać* + *-ć* | *zak\|aza\|ć* | *zak* does not directly correspond to the prefix *za-*. |

| Form | Morphological / derivational level | Example segmentation | Main interpretation |
|---|---|---|---|
| *zakaz* | Nominal form related to the *zakazać* family | *zak\|az* | The tokenizer stabilizes recurring surface fragments, not the derivational relation itself. |
| *zakazany* | *zakaz + -any* | *zak\|az\|any* | The final fragment *any* is stabilized, but this does not equal full grammatical analysis. |
| *pokazujemy* | 1st person plural, present tense | *pok\|az\|uj\|emy* | *emy* may coincide with a personal ending position, but it does not by itself anchor the full verbal form. |

***Table 5. Nasal vowels and diacritics***

### 3.6. Summary of Results

The collected data lead to five main observations.

First, BPE tokenization in Polish does not systematically move from written form to the segmentation-phonemic or phonetic-realizational level. In words such as *chleb*, *lód*, *nóż*, *prośba*, and *ławka*, the tokenizers stabilize orthographic representation, but they do not realize the relation between spelling and pronunciation or context-dependent phonetic rules. The example *jabłko* additionally shows that tokenization stabilizes written form also in cases where phonetic realization may involve a consonant cluster, pronunciation variability, or reduction. This is not, however, a "pronunciation error" on the part of the tokenizer, but a consequence of the fact that BPE operates at the graphemic-frequency level.

Second, tokenization is not a simple letter-based division. It stabilizes frequent graphemic sequences such as *zię*, *ich*, *ani*, *aza*, *ć*, *emy*, or *uj*. It is therefore most accurately described as graphemic-frequency segmentation. BPE does not divide a word simply letter by letter; it stabilizes those fragments of written form that reach sufficient frequency in the corpus. This result is consistent with the basic logic of BPE as an algorithm of compression and subword segmentation, but at the same time it reveals its limitations in an inflectional language.

Third, BPE may locally produce segments convergent with syllables, logotomes, or morphologically interpretable fragments. This convergence does not, however, mean systematic recognition of language structure. Rocławski's segmentation-flexional forms remain, in this article, a diagnostic pattern rather than a description of how BPE operates. If a token coincides with a syllable, logotome, or morphological fragment, this is a local convergence that requires interpretation, not evidence that the tokenizer represents that level of language.

Fourth, BPE stabilizes statistical traces of grammar, but it does not guarantee grammatical form anchoring. Final fragments such as *-y*, *-ich*, *-ni*, or *-ć* may appear as stable tokens or parts of tokens, but this does not mean that the full grammatical function within the inflectional system is preserved. In *ustanawiamy*, final *-y* is stabilized, but the full verbal form requires reference to person, number, tense, mood, and aspect. In *ogólnoludzkich*, *-ich* is stabilized, but the full interpretation of the form requires reference to case, number, and gender. BPE may therefore preserve a fragment of an exponent, but not the whole grammatical form.

Fifth, tokenization alone is not sufficient for a stable representation of Polish. Grammatical form must be anchored in the inflectional system and then in sentence structure. The verb is especially important here, because it organizes relations between the participants of an event. The analysis of forms such as *ustanawiamy*, *przekazać*, *pokazujemy*, and *ogólnoludzkich* shows that token boundaries may cut through places important for inflection, derivation, and sentence-level relations.

The collected results therefore shift the research question. It is not enough to ask whether BPE "correctly divides" a Polish word. One must ask at what level a given segmentation operates and whether it allows the relation between written form, segmental structure, inflectional form, place in the sentence, and the position of the speaking subject to be maintained. The first five groups of results concern tokenization data. Only the discussion extends this problem to another level: the grammatical "I" as the position of the speaking subject, which in Polish may also be expressed through form.

## 4. Discussion

The results of the analysis show that the problem of Polish tokenization cannot be reduced to the question of whether a given word is divided into few or many tokens. The more important question is the level of linguistic description on which a given segmentation operates, and which relations are preserved, cut through, or masked by it.

BPE operates at the graphemic-frequency level: it stabilizes frequent sequences of written form. Polish, however, requires the maintenance of relations between orthographic representation, the segmentation-phonemic level, phonetic realization, morpho-inflectional structure, sentence pattern, and the position of the speaking subject. It is precisely the divergence between the level at which the tokenizer operates and the levels at which the Polish linguistic system is organized that constitutes the main object of discussion.

The results do not prove that BPE "understands" or "does not understand" Polish in a psycholinguistic sense. They show, rather, that tokenization may produce local convergences with units of linguistic description — a syllable, a logotome, a morphological fragment, an inflectional ending — without preserving their systemic function. In this sense, BPE may stabilize statistical traces of grammar, but it does not guarantee grammatical form anchoring.

### 4.1. BPE Stabilizes Written Form, Not Language as a System

The first conclusion from the analysis concerns the basic level at which tokenization operates. BPE does not operate on sounds, phonemes, syllables, logotomes, morphemes, or grammatical forms. It operates on sequences of written form and their frequency. Its units may locally coincide with linguistic units, but such coincidence does not mean that the levels are identical.

This distinction is especially important in Polish, where the same written sequence may belong to different levels of description at the same time. A fragment such as *-y* may be part of an inflectional exponent, but it may also simply be a frequent final graphemic sequence. A fragment such as *-ich* may be associated with an adjectival form, but it does not in itself preserve case, number, and gender. A fragment such as *zię* may look close to a segment significant for Polish phonology or orthography, but in BPE it remains primarily a stabilized fragment of written form.

Therefore, the local stability of a token must not be confused with linguistic anchoring. A token may look like a syllable, a logotome, a morpheme, or an ending, but its status in BPE results from frequency-based compression, not from linguistic analysis. This is why BPE can be useful as a statistical tool and at the same time insufficient as a representation of the Polish language system.

From this perspective, the question is not whether BPE "makes mistakes" in a traditional linguistic sense. Rather, the question is what BPE does not represent because of the level on which it operates. It does not represent the transition from spelling to phonetic realization, it does not represent the relation between an inflectional ending and a paradigm, and it does not represent the role of the verb in organizing a sentence. These relations must be modeled or maintained by other layers of the system.

This distinction is consistent with the technical nature of BPE as a method of compression and subword segmentation. The algorithm was designed to stabilize frequent sequences, not to reconstruct the phonology, morphology, or syntax of a language (Gage, 1994; Sennrich et al., 2016). Research on subword tokenization shows that the efficiency of such segmentation is not equivalent to its correspondence with morphological units, especially in morphologically rich languages (Bostrom & Durrett, 2020; Gutierrez-Vasques et al., 2023; Držík & Kapusta, 2026).

In the case of Polish, this means that tokenization may preserve a surface fragment of written form while losing the relation between that fragment and the language system. The examples *lód* and *nóż* show that BPE stabilizes the written sequences *ód* or *óż*, but does not move to the orthographic-phonemic relation in which *ó* represents /u/, nor to the realizational level, where final devoicing occurs. Similarly, *morze* and *może* may be phonemically equivalent, but tokenization differentiates them because of the spelling contrast *rz/ż*.

This is not a "phonetic error" of the tokenizer. A tokenizer is not a phonetic tool. The problem is that written Polish is not a transparent mapping of all phonological, phonetic, and inflectional relations. If the model receives only written form, and if its tokenization stabilizes the frequency of written form, then it also stabilizes historical and systemic orthographic difficulties as graphemic fragments. Rocławski's theory makes it possible to name these places, but this does not mean that BPE recognizes them in a linguistic sense (Rocławski, 1976, 2010).

### 4.2. Apparent Correctness of Segmentation

In the material analyzed, there were cases in which a token boundary appeared to correspond to a linguistic unit. The segment *ław|ka* may give the impression of correct segmentation because it coincides with the syllabic division *ław-ka*. Similarly, *pro|ś|ba* may look like a segmentation close to the syllabic-logotomic level. Such cases are important because they reveal the phenomenon of apparent correctness.

Apparent correctness means local agreement with one level of description while divergence remains at another level. In the case of *ławka*, agreement with the syllable does not mean agreement with the phonetic-realizational level, because *w* is devoiced to *f* before *k*. In the case of *prośba*, the local extraction of *ś* does not mean a move to phonetic realization, because before voiced *b*, *ś* is voiced to *ź*. Segmentation may therefore be partly convergent with written form or syllable structure, while remaining non-phonetic in context.

This distinction is crucial for evaluating tokenization in Polish. It is not enough to say that a token boundary "matches" some fragment of the word. One must specify which level it matches: orthographic, syllabic, logotomic, morphological, or realizational. Without such a distinction, every local convergence could be mistakenly read as evidence that the tokenizer has recognized language structure.

In this sense, Rocławski serves a diagnostic function in the article. His distinctions make it possible to see when tokenization approaches the functional divisibility of a word, and when it only stabilizes a frequent fragment of written form. The claim, therefore, is not that BPE discovers logotomes. Rather, the point is that logotomes, syllables, and segmentation-phonemic units provide a comparative grid through which one can assess which level of structure has been preserved and which has merely been simulated by the frequency of written form (Rocławski, 1976, 2010).

### 4.3. The Statistical Shadow of Grammar

The second important conclusion is that BPE may stabilize fragments resembling grammatical exponents, but it does not necessarily preserve grammatical form as a unit of the system. Segments such as *-y*, *-ich*, *-ni*, *-ć*, *aza*, *uj*, or *emy* are not random in the sense that they often recur in places important for Polish morphology and inflection. This does not mean, however, that the tokenizer recognizes a grammatical category.

For this reason, I use the concept of the **statistical shadow of grammar** in this article. It refers to a situation in which a frequent sequence of written form locally coincides with a place that is morphologically or inflectionally significant, but does not constitute evidence of full grammatical form anchoring. Understood in this way, the shadow of grammar is an effect of the graphemic-frequency level, not evidence of morphological representation.

The example *ustanawiamy* shows this problem very clearly. Tokenization may stabilize final *-y*, but the full verbal form requires recognition of the first person plural, present tense, indicative mood, and imperfective aspect. Similarly, in *ogólnoludzkich*, final *-ich* may be stable as a fragment of written form, but the full interpretation of the form requires reference to case, number, and gender. These categories belong to the inflectional system, not to the graphemic surface alone.

The statistical shadow of grammar should not be confused with morphological analysis of the form. Morphological analyzers for Polish, such as Morfeusz, assign possible lexical and morphosyntactic interpretations to forms; BPE does not do this. The tokenizer does not indicate a lemma, grammatical category, or inflectional paradigm. It stabilizes a sequence of written form that may locally resemble a grammatical element.

The concept of the statistical shadow of grammar makes it possible to avoid two simplifications. The first would be to claim that if a token coincides with an ending or a morphological fragment, then the model has recognized grammar. The second would be to claim that if tokenization cuts through a form, then it preserves nothing linguistically relevant. The results show an intermediate situation: BPE may preserve traces of the system, but these traces do not yet constitute anchoring of the form.

These two concepts describe two sides of the same operation: the pseudo-inverse of lemmatization names what BPE does to the form — fragmenting it below the morpheme by frequency — while the statistical shadow of grammar names what remains afterward: a surface trace that coincides with a grammatical position without anchoring the form.

### 4.4. Rocławski as a Diagnostic Tool, Not a Description of BPE

The results of the analysis make it possible to clarify the role of Rocławski's theory in the study of tokenization. Rocławski does not provide a description of how BPE operates. It should therefore not be said that the tokenizer "discovers" sounds, logotomes, or segmentation-flexional forms in Rocławski's sense. BPE operates at the graphemic-frequency level, whereas Rocławski's theory makes it possible to build a diagnostic grid for assessing which level of linguistic description token boundaries locally coincide with.

This distinction is crucial. If a token segment coincides with a syllable, a logotome, or a morphologically interpretable fragment, this does not automatically mean that the tokenizer possesses a representation of that level. It only means that a frequent sequence of written form, in that particular place, coincides with a unit relevant for the description of Polish. This is precisely why the analysis requires separating the graphemic-frequency, segmentation-phonemic, phonetic-realizational, and morpho-inflectional levels.

Rocławski's distinction between basic and non-basic letters is especially important here. It makes it possible to name the places where orthographic representation is not transparent in relation to phonemic structure or phonetic realization. From the perspective of BPE, these are simply frequent graphemic configurations. From the perspective of language, they are sites of orthographic and segmentational difficulty. A tokenizer may stabilize them as stable tokens, but this does not mean that it recognizes the relation between written form and the segmentation-phonemic level (Rocławski, 1976, 2010).

The examples *morze/może*, *lód/nóż*, and *ząb* show this difference especially clearly. In *morze/może*, the same segmentation-phonemic level may be represented by different spellings. In *lód* and *nóż*, the spelling *ó* corresponds to /u/, but tokenization stabilizes the orthographic sequences *ód* and *óż*. In *ząb*, the grapheme *ą* functions as an element of written form, whereas phonetic realization depends on context. In each of these cases, BPE remains on the surface of written form, while Rocławski makes it possible to indicate where that written form ceases to be transparent.

Rocławski is therefore used in this article as a comparative and diagnostic tool. His theory makes it possible to ask whether token boundaries approach the functional divisibility of the Polish word, or whether they remain a statistical fragmentation of written form. It does not serve to attribute phonological or logotomic competence to the tokenizer. This distinction is important because it protects the analysis against a category-level error: BPE does not operate on the same material as Rocławski's phonemic-didactic description.

In this sense, Rocławski's greatest value for the analysis of tokenization lies not in the possibility of directly "building" his system into BPE, but in the fact that it provides a precise language for the places where Polish spelling, segmentation, and pronunciation diverge. This makes it possible to see that tokenization is not a neutral preparation of text, but the first point at which an AI system encounters the orthographic surface of Polish — a surface that is historically and structurally heterogeneous.

### 4.5. The Form Must Enter the Sentence

The next conclusion concerns the limits of tokenization analysis itself. Even if a tokenizer were better at stabilizing sublexical units, this would not solve the entire problem of Polish. Grammatical form is not self-sufficient at the level of the word. Its full function becomes visible only in the sentence, in relation to the verb, the sentence pattern, and the valency structure.

The example *ustanawiamy* shows that final *-y* may be statistically stable, but it is not sufficient on its own for recognizing the full verbal form. The form *ustanawiamy* becomes functional only in a sentence: someone establishes something, in some mode, within a specific communicative or normative act. Similarly, *przekazać* is not merely a form ending in *-ć*. This verb opens positions for the sender, the recipient, and the content being transmitted. Token segmentation may preserve fragments of written form, but it does not replace syntactic-semantic relations.

Sentence patterns and valency are especially important here. In syntactic-semantic approaches, the verb is not merely one part of the sentence, but the center that organizes relations between the participants of an event. It determines possible argument positions and links them with morphological forms. Therefore, stable modeling of Polish requires not only a better tokenizer, but also the maintenance of relations between the verbal form, its valency, and sentence structure (Grzegorczykowa, 2007; Fillmore, 1968; Tesnière, 1959).

This point is particularly important for Polish because word order is relatively free and many grammatical relations are encoded through inflection. In a more strongly positional language, some relations may be reconstructed from word order. In Polish, linear position alone is often not enough. The model must maintain relations between forms, cases, agreement, aspect, and argument structure. Tokenization may provide input material, but it cannot replace these relations.

The concept of grammatical form anchoring proposed in this article therefore goes beyond segmentation. Anchoring means maintaining the relation between the form and the inflectional system, and then between inflection and the sentence. If the form is not embedded in sentence structure, a stable token fragment remains only a fragment of written form. In this sense, the problem of tokenization is the beginning of the problem, not its end.

In the material analyzed, BPE did not function as a phonological, phonetic, morphological, or syntactic analyzer. It functioned according to its own nature: it stabilized frequent sequences of written form. The problem arises because, in Polish, written form is not a transparent carrier of all linguistic relations. Letters, digraphs, diacritics, inflectional endings, and derivational components are not merely graphemic fragments. In many places, they point to levels that BPE itself does not recognize.

Rocławski's theory makes it possible to capture the first layer of this problem: the relation between written form, the segmental unit, the phonemic level, the syllable, and the logotome. Inflectional analysis reveals the second layer: the relation between ending, form, and the inflectional system. Sentence-valency analysis reveals the third layer: the relation between the verbal form, arguments, and sentence structure. The problem of the grammatical "I" opens a fourth layer: the relation between linguistic form and the position of the speaking subject in dialogue.

From this perspective, a stable Polish language model cannot be evaluated solely by whether it generates fluent responses or by how many tokens it uses to process text. It must be evaluated by its ability to maintain relations. Tokenization is the first stage of this work, but not its end. It may facilitate or hinder the later reconstruction of form, but it does not itself replace the system of inflection, syntax, or valency.

The most important conclusion of the discussion is therefore as follows: for Polish as an inflectional language, the quality of tokenization should be assessed not only in terms of compression, but also in terms of anchoring. A stable token is not yet a stable form. A stable form is not yet a stable sentence. A stable sentence is not yet a stable dialogical "I".

## 5. Limitations of the Study

The study is qualitative, exploratory, and diagnostic in character. The selection of material was purposive rather than random. The aim was not to estimate the frequency of particular types of segmentation across the entire Polish language, but to capture especially diagnostic sites: those where graphemic-frequency tokenization diverges from orthographic representation, the segmentation-phonemic level, phonetic realization, morpho-inflectional structure, or sentential relation.

The first limitation is therefore the scope of the material. The analyzed examples include a children's text, selected official forms, a word family, phonologically diagnostic words, and examples with nasal vowels and diacritics. This material makes it possible to identify mechanisms and interpretive tensions, but it is not sufficient for building a full benchmark of Polish tokenization. Such a benchmark would need to include a larger corpus, diversified in terms of genre, style, and frequency.

The second limitation is the dependence of results on the tokenizer version and testing environment. Tokenization may differ across model families, library versions, interfaces, and the date of testing. For this reason, a replicable version of the study must provide full technical data: the model name, tokenizer name, library version, date of testing, input text, token list, and token identifiers. The labels used in the article — Bielik/APT4 Tokenizer, OpenAI-current Tokenizer, and OpenAI-legacy Tokenizer — are descriptive and serve to distinguish the observed tokenization environments; they do not replace official architectural documentation.

The third limitation is the expert character of the classification. Assessing whether a given segment is convergent with a syllable, logotome, morphological fragment, inflectional exponent, or orthographic difficulty requires linguistic expertise. In a future version of the study, it would be advisable to use a second coder, a table of classification decisions, and a procedure for resolving ambiguous cases.

The fourth limitation is the separation between tokenization and model representation. Token analysis shows what units enter the model as input, but it does not directly show what representations are formed in the hidden layers or how the model reconstructs morphological and syntactic relations in later processing. The findings therefore concern the quality of input segmentation and its potential consequences, not the full linguistic competence of the model.

The fifth limitation is the separation between the segmentation-phonemic level and the phonetic-realizational level. Examples such as *ząb*, *ręka*, *kąt*, *kąpiel*, *lód*, *nóż*, *prośba*, and *ławka* require careful distinction between orthographic representation, the segmentational level, and actual or normatively described phonetic realization. In the article, phonetic realization functions as a reference level, but it is not equated either with the operation of BPE or with Rocławski's theory.

The sixth limitation is the status of the grammatical "I". It is not a classical tokenization test, but an interpretive extension of the problem of grammatical form anchoring to the level of utterance generation by a language model. Its systematic analysis would require a separate dialogical study, including the stability of first-person, gendered, and personal forms in longer human–model interactions.

The seventh limitation is the scope of the architectural hypothesis. The proposed layers — sublexical, morpho-inflectional, sentence-valency, and subject-dialogical — are neither a ready implementation nor a decisive result. They constitute a design framework and a program for further research. At this stage, one may say that the qualitative analysis reveals the need for such a multilayered approach, but it does not yet determine how it should be technically implemented.

The eighth limitation is the lack of direct measurement of the effect of tokenization on generation quality. The article identifies places where tokenization may hinder the preservation of grammatical form, but it does not yet systematically test whether particular types of segmentation lead to specific generative errors. Such a step would require a separate experiment combining tokenization analysis with evaluation of model responses in inflectional, syntactic, and dialogical tasks.

These limitations do not undermine the central conclusion of the article. Rather, they show that the analysis presented here should be treated as a starting point for a broader research program: one that connects tokenization analysis, the morphology of Polish, sentence patterns, valency, and the stability of the grammatical "I" in dialogue.

## 6. Conclusions

The analysis presented in this article shows that BPE tokenization in Polish cannot be assessed solely by the number of tokens assigned to a word or text. In an inflectional language, the more important question is what happens to the relation between written form, segmental structure, grammatical form, the sentence, and the position of the speaking subject.

BPE operates at the graphemic-frequency level. It stabilizes frequent sequences of written form, not units of language as such. A token may locally coincide with a syllable, a logotome, a morphological fragment, or part of an inflectional exponent, but such convergence does not yet mean that the tokenizer recognizes that level of language structure. This is especially important in Polish, where orthographic representation, the segmentation-phonemic level, phonetic realization, inflection, and sentential relations are not transparent to one another.

Rocławski's theory proved useful as a diagnostic tool, but not as a description of how BPE operates. It makes it possible to indicate places where written form, the segmental unit, syllable, logotome, and phonetic realization diverge. This allows us to see that tokenization may appear to approach the structure of language while still remaining segmentation of written form. BPE does not tokenize the phonology of Polish; it tokenizes its orthographic surface, together with historically and systemically stabilized difficulties of spelling.

The analysis of forms such as *ustanawiamy*, *ogólnoludzkich*, *przekazać*, and *pokazujemy* shows that a stable token fragment is not equivalent to an anchored grammatical form. An ending, or part of an ending, may be present in tokenization, but the full form requires reference to the inflectional system: person, number, gender, case, tense, mood, and aspect. For this reason, the article distinguishes the statistical shadow of grammar from grammatical form anchoring.

The next step is the sentence. In Polish, grammatical form does not function in isolation. It enters sentence patterns and valency structures organized primarily by the verb. Therefore, stable modeling of Polish requires not only economical tokenization, but also the maintenance of relations between the form, the inflectional paradigm, the verb, and the argument structure of the sentence.

The problem extends further to the grammatical "I". In Polish, the speaking subject may be anchored in the form itself, without a separate pronoun. Forms such as *poszłam*, *zrobiłam*, or *byłam* do not merely describe an action or state; they also establish the position of the speaking subject. A language model, however, does not possess its own stable grammatical "I". It reconstructs this position contextually and statistically, which may lead to mirroring or shifting grammatical gender in dialogue.

The main conclusion is therefore that the problem of Polish AI is not reducible to the number of tokens. Reducing token count may improve efficiency, but it does not by itself preserve grammatical form, sentence relations, or the position of the speaking subject. For Polish, a more adequate approach requires several layers: sublexical stabilization, morpho-inflectional anchoring, sentence-valency representation, and subject-dialogical stability.

In this sense, tokenization is the first threshold at which an AI system encounters Polish. If this threshold remains purely graphemic and frequency-based, the later layers must compensate for the loss or fragmentation of relations that are essential for an inflectional language. The task is therefore not to replace BPE with a simple linguistic segmentation, but to recognize that tokenization must be evaluated in relation to the levels of language that Polish requires.

A stable Polish language model should not only process Polish more cheaply. It should preserve the relations that make Polish grammatical: between spelling and segmental structure, between form and inflection, between verb and sentence, and between sentence and the speaking subject. This is the central hypothesis that follows from the analysis and the starting point for further research.

Appendix A. Diagnostic Register: Reference Levels and Interpretation of Tokenization

his appendix presents the full register of diagnostic examples used in the analysis. It is not a table intended for the automatic evaluation of tokenization as correct or incorrect. The purpose of the appendix is to separate the reference levels: orthographic representation, the segmentation-phonemic level, the phonetic-realizational level, the syllabic level, the logotomic level, the morpho-inflectional level, and the interpretation of tokenization.

The table follows the principle that **tokenization is fixed, while interpretation is revisable**. This means that the recorded token segmentation remains raw data, whereas the linguistic comment specifies which level of description a given segmentation is convergent with or divergent from.

Appendices

**Appendix A1. Phonologically and Orthographically Diagnostic Words**

| Word | Reference levels | Main diagnostic function | Interpretive comment |
|---|---|---|---|
| *jabłko* | Orthographic form with the consonant cluster *b-ł-k*; syllabically *jabł-ko*. The example requires realizational caution. | Relation between spelling, consonant cluster, and possible pronunciation variability. | This example does not belong to the group of nasal vowels. It shows that tokenization stabilizes orthographic representation rather than the phonetic-realizational level. |
| *chleb* | *ch* as the spelling of /x/; final *b* is devoiced to [p] at the realizational level; monosyllabic form. | Relation between the digraph *ch* and final devoicing. | Tokenization such as *ch*\|*le*\|*b* or *ch*\|*leb* preserves spelling, not phonetic realization. |
| *ząb* | Orthographic *ą* as a grapheme; realizationally, nasality is decomposed before a labial consonant and final *b* is devoiced; provisionally *z-o-m-p*. | Separation of written form from phonetic realization. | There is no contradiction between *ą* as a unit of written form and *z-o-m-p* as a provisional realizational description. These are different levels. |
| *lód* | *ó* as the spelling of /u/; final *d* is devoiced to [t]; provisionally *l-u-t*. | *ó* as orthographic representation of /u/, not a process of “ó → u”. | Tokenization such as *l*\|*ód* stabilizes the spelling *ód*. |
| *nóż* | *ó* as the spelling of /u/; final *ż* is devoiced to [sz]; provisionally *n-u-sz*. | Relation between orthographic form and final devoicing. | Tokenization stabilizes written form and does not represent final devoicing. |
| *lekarz* | *rz* as the spelling of /ż/; final pronunciation requires caution. | Relation between the grapheme sequence *rz* and its phonemic value. | Tokenization preserves the written sequence rather than reducing *rz* to its phonemic value. |

| Word | Reference levels | Main diagnostic function | Interpretive comment |
|---|---|---|---|
| *morze* | Spelling with *rz*; segmentation-phonemic level shared with *może*: *m-o-ż-e*. | Control pair for *rz* / *ż*. | BPE differentiates spelling, not phonemic equivalence. |
| *może* | Spelling with *ż*; segmentation-phonemic level shared with *morze*: *m-o-ż-e*. | Control pair for *rz* / *ż*. | BPE preserves the graphemic difference between *rz* and *ż*. |
| *jesień* | Orthographic form with *ś* and final *ń*; syllabically *je-sień*. | Relation between spelling, palatal consonants, and syllabic structure. | Tokenization may locally approach syllabic division, but it remains based on written form. |
| *prośba* | Spelling with *ś*; realizationally, *ś* is voiced to [ź] before *b*. | Relation between written *ś* and contextual voicing. | Tokenization such as *pro\|ś\|ba* is mixed: locally convergent with written/syllabic structure, but not with phonetic realization. |
| *ławka* | Syllabically *ław-ka*; realizationally, *w* is devoiced to [f] before *k*. | Difference between syllabic convergence and phonetic realization. | Tokenization such as *ław\|ka* may coincide with syllabic division, but it does not represent the realizational form *ł-a-f-k-a*. |

**Note.** This appendix separates reference levels used in the interpretation of diagnostic words. The table does not classify tokenization as simply correct or incorrect. It indicates which level of Polish linguistic description is relevant for each example: orthographic representation, segmentation-phonemic relation, syllabic/logotomic structure, or phonetic-realizational form.

**Appendix A2. Official and Morphologically Complex Forms**

| Word | Reference levels | Main diagnostic function | Interpretive comment |
|---|---|---|---|
| *Rzeczypospolitej* | Syllabically: *rze-czy-pos-po-li-tej*. Historically and derivationally complex form. | Test of long-form segmentation and the relation between syllabic, logotomic, and graphemic-frequency levels. | The number of tokens alone does not determine segmentation quality. It is necessary to ask which level is preserved: syllabic, logotomic, morphological, or only graphemic-frequency. |
| *zobowiązani* | Syllabically: *zo-bo-wią-za-ni*. Inflected participial/adjectival form; final *-ani* as a surface fragment. | Test of whether final *-ani* stabilizes as a written fragment or as part of grammatical form. | Final *-ani* may stabilize as a frequent fragment, but this does not mean that the tokenizer preserves syllabic division or full grammatical anchoring. |
| *ustanawiamy* | Syllabically: *u-sta-na-wia-my*. Verbal form: 1st person plural, present tense, indicative mood, imperfective aspect; full personal ending *-my*. | Test of the relation between tokenization and personal verbal ending. | Final *-y* may stabilize, but the full personal ending *-my* is not preserved as one segment. Stable tokenization is therefore not equivalent to grammatical form anchoring. |
| *ogólnoludzkich* | Morphologically: *ogólno-ludzk-ich*. Inflected adjectival form anchored in declension, number, gender, and case. | Test of whether final *-ich* is preserved as a grammatical exponent or only as a frequent written fragment. | Final *-ich* may be stable as a surface fragment, but this does not mean that the tokenizer preserves the full grammatical interpretation of the form. |
| *przekazać* | Morphologically: *prze-* + *kazać* + *-ć*. Infinitive form with prefix, base, and infinitive marker. | Test of prefix and base preservation within a derived verbal form. | The prefix *prze-* and the base *kaza-* are cut through in observed segmentations. Final *-ć* may stabilize, but only as a fragment of the grammatical form. |
| *wdzięczni* | Graphemic-frequency level: written sequences *dzię*, *zię*, *cz*, *ni*; final *-ni*. | Test of segmentation involving Polish diacritics and frequent written sequences. | Segmentation depends on the frequency of written form. Final *-ni* may stabilize, but this does not amount to full morpho-inflectional anchoring. |

**Note.** This appendix identifies the reference levels used for interpreting selected official and morphologically complex forms. The examples show that BPE may stabilize fragments such as *-ani*, *-y*, *-ich*, *-ć*, or *-ni*, but such stabilization should be interpreted as a graphemic-frequency effect unless the form can be related to the inflectional system and to its grammatical function.

**Appendix A3. The *kazać / pokazać / zakazać* Word Family**

| Form | Morphological / derivational level | Example segmentation | Interpretive comment |
|---|---|---|---|
| *przekazać* | *prze-* + *kazać* + *-ć* | *prz\|ek\|aza\|ć* | The prefix *prze-* and the base *kaza-* are not preserved as wholes. |
| *kazać* | *kaza-* + *-ć* | *k\|aza\|ć* | The fragment *aza* stabilizes as a surface fragment, but not as the full base *kaza-*. |
| *pokazać* | *po-* + *kazać* + *-ć* | *pok\|aza\|ć* | The fragment *pok* does not directly correspond to the prefix *po-*. |
| *zakazać* | *za-* + *kazać* + *-ć* | *zak\|aza\|ć* | The fragment *zak* does not directly correspond to the prefix *za-*. |
| *zakaz* | Nominal form related to the *zakazać* family. | *zak\|az* | The tokenizer stabilizes recurring surface fragments, not the derivational relation itself. |
| *zakazany* | *zakaz* + *-any* | *zak\|az\|any* | The final fragment *any* is stabilized, but this does not amount to full grammatical analysis. |

| Form | Morphological / derivational level | Example segmentation | Interpretive comment |
|---|---|---|---|
| *pokazujemy* | Verbal form: 1st person plural, present tense. | *pok\|az\|uj\|emy* | The fragment *emy* may coincide with the position of a personal ending, but it does not by itse |

**Note.** This appendix shows that, within the *kazać / pokazać / zakazać* word family, tokenization primarily stabilizes recurring written fragments such as *aza*, *az*, *zak*, *pok*, *ć*, *any*, *uj*, and *emy*. This should not be interpreted as evidence that the tokenizer preserves the morphological family as a system of derivational and inflectional relations.

**Appendix A4. Nasal Vowels and Diacritics**

| Word | Orthographic representation | Phonetic-realizational level | Example tokenization | Interpretive comment |
|---|---|---|---|---|
| *ręka* | *ręka* | *ę* before *k*: realization close to *renka*; oral vowel + velar nasal consonant. | *rę\|ka* | Tokenization stabilizes the written sign *ę*, not phonetic realization. |
| *kąt* | *kąt* | *ą* before *t*: realization close to *kont*. | *ką\|t* | Tokenization stabilizes the grapheme *ą*. |
| *kąpiel* | *kąpiel* | *ą* before *p*: realization close to *kompiel*. | *ką\|piel* | Written *ą* is not equivalent to one stable phonetic realization. |
| *wąski* | *wąski* | *ą* before the fricative *s*; in Rocławski's approach, realization as nasal *õ*, IPA /võski/. | *wą\|ski* | This is not the same mechanism as before stop consonants. BPE stabilizes the written sequence *wą*. |
| *koń* | *koń* | Final *ń* as a palatal consonant sign. | *koń* | The whole form may be stable as one token. |
| *pień* | *pień* | Final *ń*; monosyllabic form. | *pie\|ń* | Monosyllabicity does not guarantee one token. |
| *dzień* | *dzień* | Frequent whole word with final *ń*. | *dzień* | The whole form is stable. |
| *cień* | *cień* | Frequent whole word with final *ń*. | *cień* | The whole form is stable. |

**Interpretive note.**

The examples in this appendix show that BPE stabilizes diacritics and frequent written sequences, but does not move to context-dependent phonetic realization. In words such as *ręka*, *kąt*, and *kąpiel*, the orthographic signs *ę* and *ą* do not correspond to one fixed phonetic realization. In *wąski*, the realization before a fricative requires separate treatment. The tokenization data therefore confirm the need to distinguish orthographic representation, the segmentation-phonemic level, and phonetic realization.

**Appendix A5. Interpretive Categories Used in the Register**

| Category | Working definition | Examples |
|---|---|---|
| Graphemic-frequency segmentation | Segmentation resulting from the frequency of written sequences, not from phonological or morphological analysis. | *zię*, *aza*, *emy*, *uj* |
| Syllabically convergent segmentation | The token boundary partially or fully coincides with a syllable boundary. | *ław\|ka* |
| Segmentation convergent with a logotome | The segment can be compared with a functional word component in Rocławski's sense, without claiming that BPE recognizes it. | *pro* in *pro\|ś\|ba* |
| Morphologically interpretable segmentation | The token boundary partially coincides with a prefix, base, suffix, or ending. | *-ć* in *przekazać*, *-ich* in *ogólnoludzkich* |
| Mixed segmentation | Tokenization shows local convergence with more than one level of description. | *pro\|ś\|ba* |
| Contextually non-phonetic segmentation | Tokenization does not account for context-dependent phonetic realization. | *ławka*, *prośba*, *lód*, *nóż* |
| Stabilization of orthographic difficulty | Tokenization stabilizes the written form of a place where the grapheme is not transparent in relation to the phoneme or pronunciation. | *morze/może*, *lód*, *nóż*, *ząb* |
| Fragmentation of a grammatical exponent | Tokenization stabilizes part of an exponent, but not the full grammatical form. | *ustanawiamy* → *-y* instead of full *-my* |
| Atomization / lack of internal segmentation | The whole word is represented as a single token, which does not imply representation of its internal structure. | *koń*, *dzień*, *cień* |
| Statistical shadow of grammar | A frequent written sequence repeatedly coincides with a morphologically or inflectionally relevant position, but does not prove recognition of a grammatical category. | *-y*, *-ich*, *-ć*, *aza*, *emy* |
| Instability of the grammatical "I" | Dialogical level, not tokenization data: the model shifts or mirrors gendered/personal forms. | *zrobiłam/zrobiłem*, *byłam/byłem* |

**Appendix B. Register of Observed Token Segmentations**

Appendix B presents the observed token segmentations used in the analysis. The register includes only the data available in the working material: visual segmentations, token counts where they were recorded, and basic test metadata. It does not include full token identifiers, because they were not available for all examples.

The appendix has an auxiliary and control function. A fully replicable version of the study should be supplemented with token identifiers, full token lists, versions of the tokenization libraries, and an exact specification of the tools used in the test.

**Author of the study:** Elżbieta Dawidek
**Period of observation and data collection:** January–April 2026
**Main test dates:** 22–26 April 2026
**Working environment:** Windows 11
**Type of data:** observed token segmentations and token counts available in the working material
**Interpretive principle:** tokenization is fixed; interpretation is revisable

**Appendix B1. Diagnostic Words**

| Word | Bielik/APT4 Tokenizer | OpenAI-current Tokenizer | OpenAI-legacy Tokenizer | Notes |
|---|---|---|---|---|
| *jabłko* | *j\|ab\|ł\|ko* | *jab\|ł\|ko* | *jab\|ł\|ko* | The example requires realizational caution; stabilization of written form. |
| *chleb* | *ch\|le\|b* | *ch\|leb* | *ch\|le\|b* | Written form is preserved; no transition to phonetic realization. |
| *ząb* | *z\|ą\|b* | *zą\|b* | *zą\|b* | The written sign *ą* must be distinguished from the realizational description *z-o-m-p*. |
| *lód* | *l\|ód* | *l\|ód* | *l\|ód* | *ó* as the spelling of /u/, not a process of "ó → u". |
| *nóż* | *n\|ó\|ż* | *n\|óż* | *n\|óż* | *ó* as the spelling of /u/; stabilization of written form. |
| *lekarz* | *le\|kar\|z* | *le\|kar\|z* | *le\|kar\|z* | The spelling *rz* remains treated graphemically. |
| *morze* | *mor\|ze* | *mor\|ze* | *mor\|ze* | Control pair with *może*. |
| *może* | *mo\|że* | *mo\|że* | *mo\|że* | Control pair with *morze*. |
| *prośba* | *pro\|ś\|ba* | *pro\|ś\|ba* | *pro\|ś\|ba* | Mixed segmentation; written *ś*, but realizationally *ś* → *ź* before *b*. |
| *ławka* | *ł\|aw\|ka* | *ław\|ka* | *ław\|ka* | Syllabic convergence does not mean phonetic convergence. |

**Appendix B2. Children's Text**

**Input text:**

*Siała baba mak, nie wiedziała jak. A dziad wiedział nie powiedział, a to było tak.*

| Tokenizer | Number of characters | Number of tokens | Observed segmentation examples | Notes |
|---|---|---|---|---|
| OpenAI-current Tokenizer | 83 | 26 | more economical segmentation | newer tokenization environment |
| OpenAI-legacy Tokenizer | 83 | 30 | higher number of tokens | older tokenization environment |
| Bielik/APT4 Tokenizer | 83 | 34 | *S\|ia\|ła*; *b\|aba*; *m\|ak*; *w\|ied\|zia\|ła*; *d\|ziad*; *pow\|ied\|ział*; *by\|ło* | examples show internal segmentation of inflectional forms |

**Appendix B3. Official and Morphologically Complex Forms**

| Form | Bielik/APT4 Tokenizer | OpenAI-current Tokenizer | OpenAI-legacy Tokenizer | Notes |
|---|---|---|---|---|
| *Rzeczypospolitej* | *R\|zec\|zy\|pos\|pol\|ite\|j* | *Rzeczy\|pos\|polite\|j* | *R\|z\|eczy\|pos\|polite\|j* | A similar number of tokens may hide different segmentation quality. |
| *zobowiązani* | *z\|ob\|ow\|ią\|z\|ani* | *z\|obowią\|z\|ani* / requires cautious treatment | *z\|ob\|ow\|ią\|z\|ani* | OpenAI-current: 5 tokens; OpenAI-legacy: 6 tokens; final *-ani* stabilizes in all observed segmentations. |
| *ustanawiamy* | *u\|stan\|aw\|iam\|y* | *ustan\|aw\|iam\|y* | *ust\|an\|aw\|iam\|y* | The full personal ending *-my* is not preserved as one segment. |
| *ogólnoludzkich* | *og\|ó\|ln\|ol\|ud\|zk\|ich* | *ogól\|no\|lud\|zk\|ich* | *ogól\|no\|lud\|zk\|ich* | Final *-ich* is stable as a fragment. |
| *przekazać* | *prz\|ek\|aza\|ć* | *prz\|ekaza\|ć* | *prz\|ekaza\|ć* | The prefix *prze-* is cut through. |
| *wdzięczni* | *w\|d\|zi\|ę\|cz\|ni* | *wdzię\|cz\|ni* | *wd\|zię\|cz\|ni* | Segmentation depends on the frequency of written form. |

**Appendix B4. The *kazać / pokazać / zakazać* Word Family**

| Form | Bielik/APT4 Tokenizer | Notes |
|---|---|---|
| *przekazać* | *prz\|ek\|aza\|ć* | The prefix *prze-* and the base *kaza-* are not preserved as wholes. |
| *kazać* | *k\|aza\|ć* | *aza* appears as a surface fragment. |

| Form | Bielik/APT4 Tokenizer | Notes |
|---|---|---|
| *pokazać* | *pok\|aza\|ć* | *pok* does not directly correspond to the prefix *po-*. |
| *zakazać* | *zak\|aza\|ć* | *zak* does not directly correspond to the prefix *za-*. |
| *zakaz* | *zak\|az* | Stabilization of surface fragments. |
| *zakazany* | *zak\|az\|any* | *any* appears as a final fragment. |
| *pokazujemy* | *pok\|az\|uj\|emy* | *emy* does not automatically mean full anchoring of the personal form. |

**Appendix B5. Nasal Vowels and Diacritics**

| Word | Observed tokenization | Notes |
|---|---|---|
| *ręka* | *rę\|ka* | The written sign *ę* is stabilized; tokenization does not reflect realizational pronunciation close to *renka*. |
| *kąt* | *ką\|t* | The grapheme *ą* is stabilized; tokenization does not reflect realizational pronunciation close to *kont*. |
| *kąpiel* | *ką\|piel* | The written sign *ą* is stabilized; tokenization does not reflect realizational pronunciation close to *kompiel*. |
| *wąski* | *wą\|ski* | The written sequence *wą* is stabilized. This example requires separate treatment because *ą* occurs before the fricative *s*. |
| *koń* | *koń* | The whole form is stable as one token. |
| *pień* | *pie\|ń* | The monosyllabic form is split; monosyllabicity does not guarantee one token. |
| *dzień* | *dzień* | The whole frequent form is stable as one token. |
| *cień* | *cień* | The whole frequent form is stable as one token. |

**Note.**

The examples in this table confirm that BPE stabilizes Polish diacritics and frequent written sequences as elements of orthographic representation. It does not move to the phonetic-realizational level. Therefore, tokenizations such as *rę|ka*, *ką|t*, *ką|piel*, and *wą|ski* should be interpreted as graphemic-frequency segmentations, not as phonetic representations.

**Replication Note for Appendix B**

Appendix B does not constitute a full replication protocol. It contains a minimal register of observed segmentations available in the working material. If full replication is required, the tests should be repeated in the same environments or in their documented equivalents, with the following information recorded:

1. the exact name of the tokenization tool;
2. the model and tokenizer version;
3. the date of testing;
4. the full token list;
5. token identifiers;
6. the method used for counting technical tokens, separators, and newline characters.

The absence of token identifiers in this appendix does not affect the qualitative analysis, because the object of the article is the relation between observed segmentation and the levels of Polish linguistic description. It does, however, limit the possibility of full technical replication without rerunning the tokenization tools.

The data presented in the appendix have a control function: the segmentations can be verified and reproduced, because the tokenizers and testing tools are publicly available or can be reused in documented environments. The results are not based on an inaccessible private corpus, but on observable tokenizer outputs applied to explicitly provided input material. This appendix provides a minimal register of observed segmentations, whereas full technical replication requires recording the token lists, token identifiers, tool versions, and test dates again.